\documentclass[11pt]{article}

\usepackage{amsmath,amssymb,graphicx}
\usepackage[numbers]{natbib}
\usepackage{geometry}
\usepackage{mathtools}
\usepackage{booktabs}
\usepackage{multirow}
\usepackage{makecell}
\usepackage{colortbl}
\usepackage{siunitx}
\usepackage{tikz}

\usepackage{hyperref}
\hypersetup{
  colorlinks=false,
  frenchlinks=false,
  pdfborder={0 0 0},
  naturalnames=false,
  hypertexnames=false,
  breaklinks
}

\include{setup_colors_jr}

\newcommand{\fpp}[1]{\num[round-precision=2,round-mode=figures]{#1}}
\newcommand{\fppp}[1]{\num[round-precision=3,round-mode=figures]{#1}}
\newcommand{\fpppp}[1]{\num[round-precision=4,round-mode=figures]{#1}}
\newcommand{\spp}[1]{\num[round-precision=2,round-mode=figures,scientific-notation=true]{#1}}
\newcommand{\sppp}[1]{\num[round-precision=3,round-mode=figures,scientific-notation=true]{#1}}

\DeclareMathOperator*{\argmin}{arg\,min}
\DeclareMathOperator{\diag}{diag}

\newcommand{\Rsq}{\ensuremath{\text{R}^2}}
\newcommand{\MSE}{\ensuremath{\text{MSE}}}
\newcommand{\var}{\ensuremath{\text{var}}}

\title{Parameter Estimation with Dense and Convolutional Neural Networks Applied to the FitzHugh--Nagumo ODE}

\author{%
Johann Rudi%
\thanks{Mathematics and Computer Science Division, Argonne National Laboratory, Lemont IL, USA}
\thanks{\texttt{jrudi@anl.gov}}%
\and
Julie Bessac\footnotemark[1]%
\and
Amanda Lenzi\footnotemark[1]%
}

\begin{document}

\maketitle

\begin{abstract}%
Machine learning algorithms have been successfully used to approximate nonlinear maps under weak assumptions on the structure and properties of the maps.
We present deep neural networks using dense and convolutional layers to solve an inverse problem, where we seek to estimate parameters of a FitzHugh--Nagumo model, which consists of a nonlinear system of ordinary differential equations (ODEs).  We employ the neural networks to approximate reconstruction maps for model parameter estimation from observational data, where the data comes from the solution of the ODE and takes the form of a time series representing dynamically spiking membrane potential of a biological neuron.
We target this dynamical model because of the computational challenges it poses in an inference setting, namely, having a highly nonlinear and nonconvex data misfit term %
and permitting only weakly informative priors on parameters.  These challenges cause traditional optimization to fail and alternative algorithms to exhibit large computational costs.
We quantify the prediction errors of model parameters obtained from the neural networks %
and investigate the effects of network architectures with and without the presence of noise in observational data.
We generalize our framework for neural network-based reconstruction maps to simultaneously estimate ODE parameters and parameters of autocorrelated observational noise.
Our results demonstrate that deep neural networks
have the potential to estimate parameters in dynamical models and stochastic processes, and they are capable of predicting parameters accurately for the FitzHugh--Nagumo model.
\end{abstract}

\section{Introduction}

We consider inverse problems with the aim of estimating parameters from given
observational data, where the parameters give rise to the data through the solution of a parametrized physical model.
Such inverse problems have in the past been solved deterministically
with techniques from optimization \citep{Tarantola05} or in a
statistical/Bayesian framework using sampling methods,
particle filters, and Kalman filters \citep{KaipioSomersalo05}.
The current work investigates new approaches to solving particular inverse problems
that exhibit computational challenges prohibiting effective use of the
established solution techniques mentioned above.
We propose to computationally learn solution operators for inverse problems based on deep neural networks because of their well-known potential of finding
generalizable nonlinear maps.  We explore
  neural network (NN)
architectures consisting of a sequence of dense, or fully connected, layers and
  convolutional neural networks (CNNs).

We target inverse problems to estimate parameters in an
  ordinary differential equation (ODE)
that models the dynamically spiking membrane potential of a (biological)
neuron.  Spiking neurons in the brain and spinal cord are typically modeled by
systems of nonlinear ODEs.
These ODE models govern electrical voltage spikes that are generated in
response to current stimuli.  The output of one such ODE has the form of a
spiking voltage time series, which can be obtained in laboratory experiments
and takes the role of observational data in our inverse problem.  These systems
of ODEs contain uncertain parameters that control the opening and closing of
ion channels of a cell membrane and consequently change voltage spike behavior.

\begin{figure}
  \centering
  \includegraphics[height=50mm]{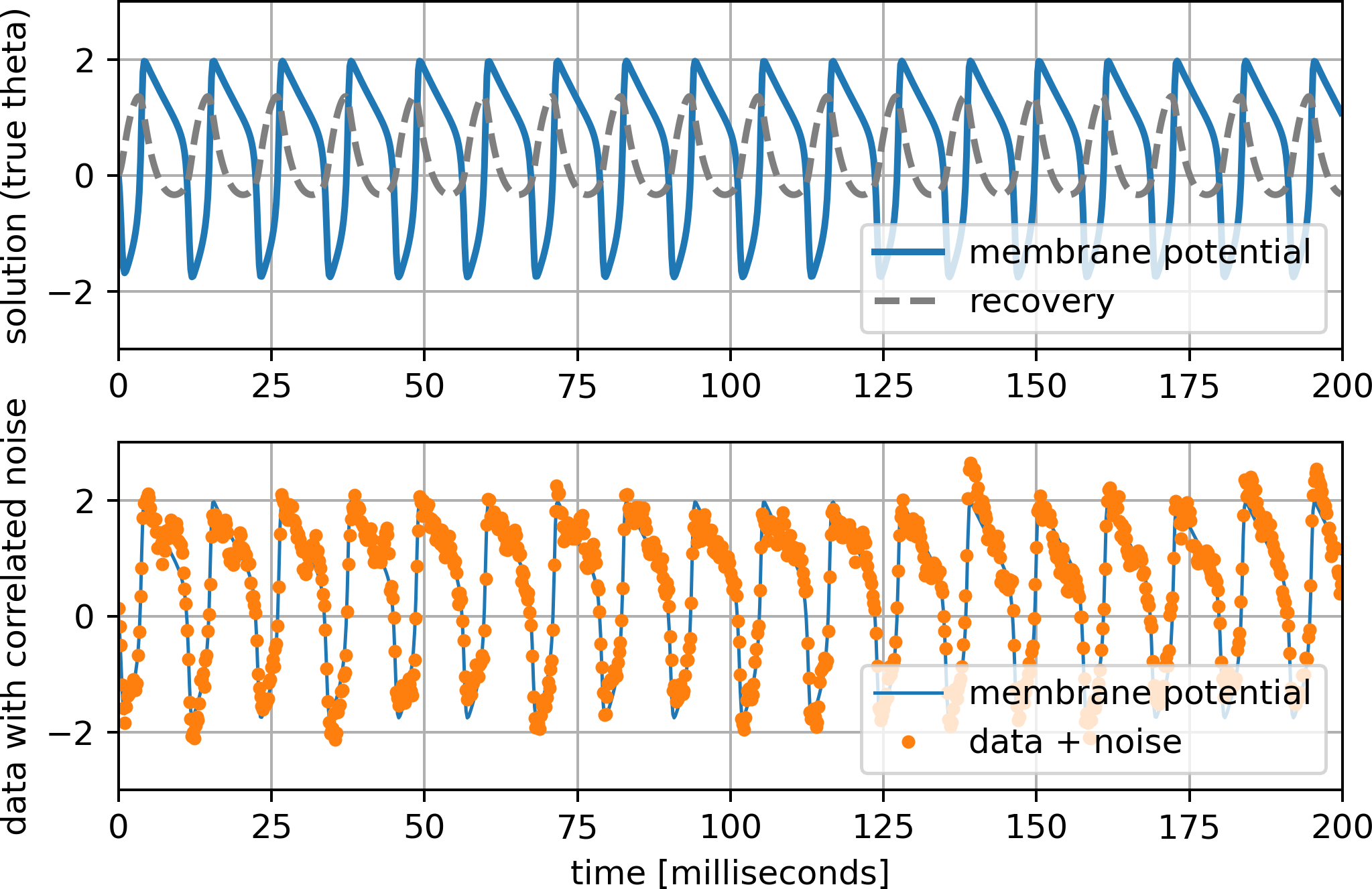}
  \hspace{2em}
  \includegraphics[height=50mm]{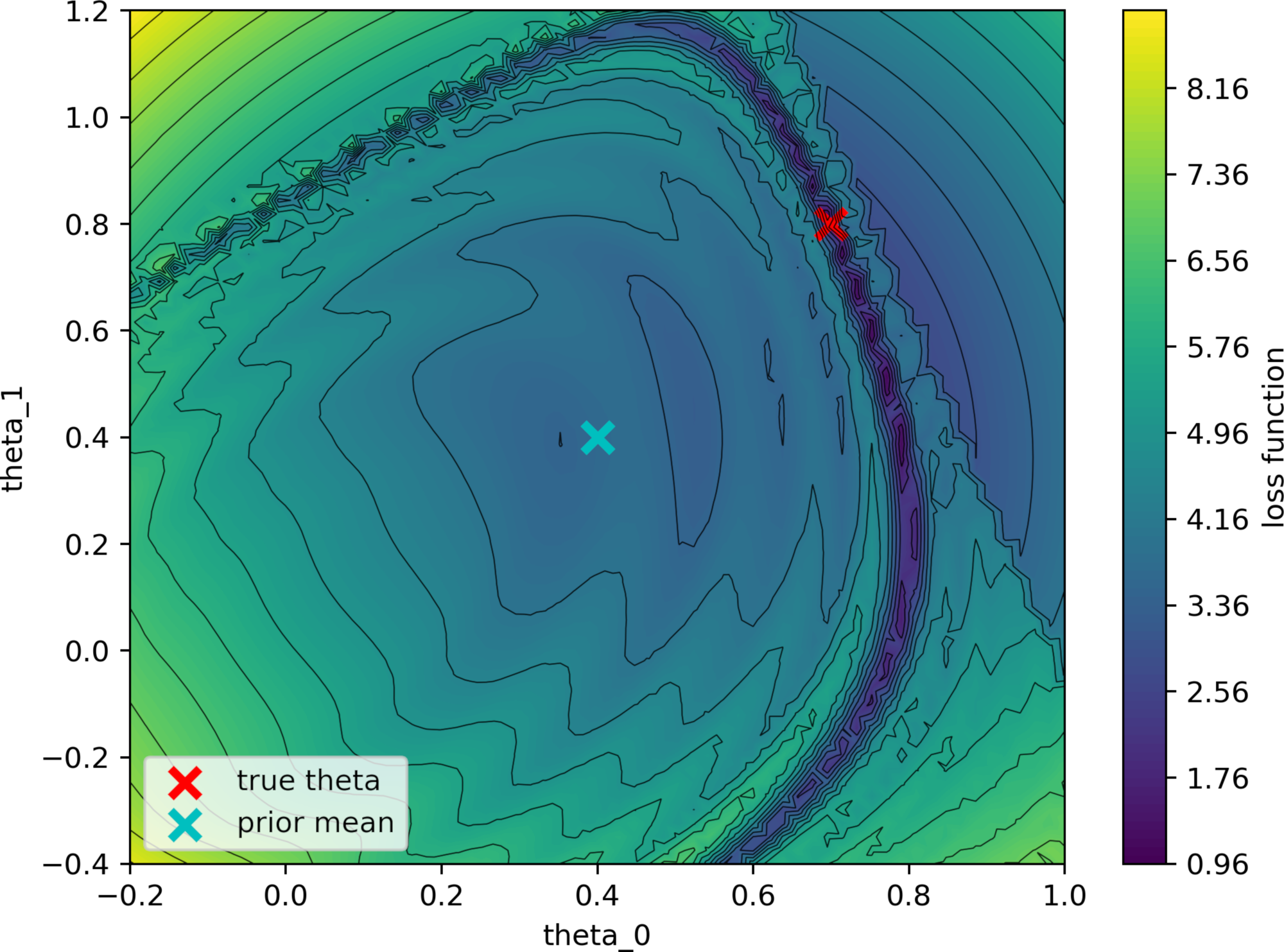}
  \caption{%
    Left, top graph: Solution of FitzHugh--Nagumo system of two ODEs (blue and gray lines)
    generated with true parameters of the inverse problem.
    Left, bottom graph: Data (orange dots) of inverse problem stemming from the membrane potential of the solution with additive correlated noise.
    Right: Manifold of highly nonlinear loss function (colors and contours), with respect to parameters $\theta_0$ and $\theta_1$, of the inverse problem governed by the FitzHugh--Nagumo model when using the data from the bottom left graph and a weakly informative (i.e., wide) prior term.
  }
  \label{fig:ode-loss}
\end{figure}

Computational challenges arise from the highly nonlinear and nonconvex loss, or
objective, function of the inverse problem (\cite{BuhrySaighiEtAl08};
see also the manifold in Figure~\ref{fig:ode-loss}, right), with sharp gradients, strong nonlinear
dependencies between parameters, and potentially multiple local minima.  The
loss cannot be sufficiently regularized by a convex additive term stemming from
a regularization or prior term, because of lack of knowledge about the range of
parameter values \citep{GutenkunstWaterfallEtAl07, PrinzBucherMarder04} and
because this would largely eliminate the information from data and model.
We therefore
explore new techniques to solve such challenging inverse problems
by fitting NN-based reconstruction maps that approximate solution operators for our inverse problem by means of mapping observational data to model parameters.

\paragraph{Contributions}

We consider neural networks with dense and convolutional layers and explore different NN architectures with a
range of layers, dense units, and CNN filters.  We quantify the capability of the networks to approximate reconstruction maps of inverse problems with several statistical metrics.
Through these metrics, we aim to provide understanding about the effects of network choice on inference performance.  The numerical results show that NNs can estimate parameters of spiking neuron models with high accuracy.

We investigate the predictive skills of NNs while changing the sizes of training sets.
Additionally, we conduct experiments on partially observed data to numerically demonstrate that CNNs, but not dense NNs, are capable of detecting underlying properties or dynamics of a time series.
Moreover, since observational data are in practice most likely corrupted by noise,
we present a noise model and analyze the influence of noise in training and/or testing data based on numerical experiments.
To go beyond estimating ODE parameters, we extend our framework for NN-based reconstruction maps to simultaneously estimate parameters of the autocorrelated noise model from noisy observational data.

\paragraph{Related work}

The literature on parameter estimation for models of neural dynamics spans across various scientific communities, approaches, and neuron models.  Here, we
highlight only closely related publications and refer to the literature within these publications for further information.
The neuron model by \cite{HodgkinHuxley52} is often used in the literature.  It consists of a nonlinear system of four ODEs.  The FitzHugh--Nagumo equations (introduced in Section~\ref{sec:forward-inverse}) simplify the Hodgkin--Huxley model to a nonlinear system of two ODEs.

Common approaches for parameter estimation in FitzHugh--Nagumo and Hodgkin--Huxley models make use of heuristics and trial and error and may consist of intricate sequences of regression steps, as summarized by \cite{BuhrySaighiEtAl08} and \cite{VanGeitEtAl08}.  The techniques employed are, for instance, simulated annealing, differential evolution, genetic algorithms, and brute-force grid search \citep{AlonsoMarder19}.  These approaches have the disadvantage of being  computationally expensive or are slow to converge.  Using gradient descent for finding the minimizer of a Hodgkin--Huxley-based inverse problem \citep{DoiOnodaKumagai02}, on the other hand, suffers from the strong nonlinearities in the objective and requires good initial guesses.
Alternative approaches estimate parameters with maximum likelihood methods and build approximations of the likelihood term \citep{DorukAbosharb19}.
Recently, Kalman filters have been utilized, for instance, by \cite{DengWangChe09};
ensemble Kalman filters are combined with reduced order models \citep{PaganiManzoniQuarteroni17}, and
augmented ensemble Kalman filters \citep{ArnoldLloyd18} are employed for periodically time-varying parameters.
In a data assimilation framework, \cite{HamiltonBerrySauer18} propose an ensemble Kalman filter for time series with large amounts of noise.
Furthermore, statistical inference in a Bayesian framework is attempted with
  approximate Bayesian computation (ABC)
\citep{DalyGavaghanEtAl15} and Monte Carlo sampling \citep{DalyGavaghanEtAl18}.
Alternatively to time series data, \cite{JoliveRauchEtAlt06} and \cite{NaudBathellierGerstner14} develop inference methods for spike time data.

While machine learning techniques generally are %
used for prediction and classification tasks in a wide range of applications, they are employed more sparsely to estimate parameters %
of mathematical models.
For instance, \cite{morshed1998} (for groundwater modeling) and \cite{dua2011} (for kinetik models) utilize NNs within their parameter estimation frameworks for systems described by partial or ordinary differential equations, they however do not consider approximating reconstruction maps.  %
\cite{parikh2020} employ generative adversarial networks to retrieve model parameters in stochastic inverse problems.
Parameter estimation approaches for neural dynamics based on NNs are proposed \citep{GoncalvesLueckmannEtAl20}, where an NN is trained to perform posterior density estimates using mixtures of Gaussian or normalizing flows.  While \cite{GoncalvesLueckmannEtAl20} show promising approximations of posterior features, their approach comes at the cost of large training sets (of order $10^5$), which is about two orders of magnitude larger than in our work.
Each training sample requires the solution of an ODE model and, if performed frequently, constitutes the major computational cost.
For such large numbers of training samples, Monte Carlo methods \citep{BallnusHugEtAl17} become preferable alternatives, especially if they could be augmented with reduced order models %
\citep{QianKramerEtAl20, RoseberryVillaEtAl20}.
Moreover, \cite{GoncalvesLueckmannEtAl20} do not analyze the effects of noise in training and/or testing data, and training set sizes are not varied.  Their network architectures are prescribed, which reflects an inductive bias \citep{CranmerBrehmerLouppe20} by knowing which network performs well for a given problem.

In statistical models, machine learning is used for direct estimation of model parameters \citep{chon1997}, where NNs are used for estimating autoregressive moving-average processes parameters.
\cite{radev2020} propose a suite of two NNs to estimate summary statistics of the data, infer parameters of a model, and return samples from the posterior distributions.
Other works focus on supplementing or complementing statistical inferential techniques with machine learning.  In particular, problems with intractable likelihood are often solved with the ABC method involving the sampling of synthetic data and summary statistics of the observations.
\cite{creel2017neural} and \cite{jiang2017learning} uses an NN to predict the parameters from artificially generated data, which are then used as an estimate of the posterior mean of an ABC procedure.

The remainder of the paper is organized as follows.  Section~\ref{sec:methods} introduces the forward problem governed by the FitzHugh--Nagumo model and the corresponding inverse problem. %
It then presents NN architectures utilized for solving the inverse problem, the generation of training and testing samples, and metrics to evaluate NN-based model parameter estimates.
Section~\ref{sec:results} discusses numerical results and quantifies prediction errors %
obtained from trained NNs.  We consider a range of network architectures, different sizes of training data, and the effects of noise.
Section~\ref{sec:conclusion} presents conclusions and open questions.  Background information is collected in the Appendix.

\section{Neural network-based reconstruction maps for inverse problems}
\label{sec:methods}

This section introduces the forward problem that is governed by the FitzHugh--Nagumo ODE and the inverse problem to estimate model parameters of the ODE from time series data.  Furthermore, we describe the NN architectures that will be used to perform the inference.  We conclude with a set of metrics used to evaluate the errors of model parameter predictions coming from NNs.

\subsection{Forward and inverse modeling of spiking neurons}
\label{sec:forward-inverse}

The FitzHugh--Nagumo \citep{Fitzhugh61, NagumoArimotoYoshizawa62} equations
describe spiking neurons via a system of two ODEs,
\begin{subequations}
\label{eq:fhn}
\begin{alignat}{1}
  \label{eq:potential}
  \frac{\mathrm{d}u}{\mathrm{d}t} &= \gamma\left(u - \frac{u^3}{3} + v + \zeta\right),\\
  \label{eq:recovery}
  \frac{\mathrm{d}v}{\mathrm{d}t} &= -\frac{1}{\gamma}\left(u - \theta_0 + \theta_1 v\right),
\end{alignat}
\end{subequations}
where the unknowns of the ODE are the membrane potential $u=u(t)$ and the
recovery variable $v=v(t)$.
$\zeta$ denotes the total membrane current and is a stimulus applied to the neuron, which we assume to be constant in time.
$\gamma$ determines the strength of damping and is assumed to be known and constant.
$\theta_0$ and $\theta_1$ are the model parameters that we consider for inference, because they govern two important characteristics of the oscillating solution of the ODE \eqref{eq:fhn}, namely spike rate and spike duration (see Appendix~\ref{sec:parameters} for more details).
ODE \eqref{eq:fhn} is
augmented with initial conditions $u(0)=u_0$, $v(0)=v_0$.  An example
solution of \eqref{eq:fhn} is shown in the top left panel of Figure~\ref{fig:ode-loss},
using inference parameters $\theta_0=0.7$, $\theta_1=0.8$, constants $\gamma=3.0$, $\zeta=-0.4$, and initial condition $u_0=v_0=0$.
While the FitzHugh--Nagumo model formulation is relatively simple compared with larger systems
of ODEs, like the Hodgkin--Huxley equations, it exhibits most of the
computational challenges when considered as the forward problem in an inference
setting, %
such as described next.

We target the inverse problem, where we are given observational data
\begin{equation}
  \label{eq:data}
  d(t) \coloneqq u_{\boldsymbol{\theta}^\ast}(t) + \eta(t),
\end{equation}
composed of $u_{\boldsymbol{\theta}^\ast}(t)$, which is the first component \eqref{eq:potential} of the solution of ODE \eqref{eq:fhn}
with unknown true parameters
  $\boldsymbol{\theta}^\ast \coloneqq (\theta_0^\ast, \theta_1^\ast)$,
and added correlated noise $\eta(t)$ (see Section~\ref{sec:results-noise} for more information about the noise model).
Our goal is to find parameters
  $\boldsymbol{\theta} \coloneqq (\theta_0, \theta_1)$
that are consistent with data $d(t)$ and model output $u_{\boldsymbol{\theta}}(t)$ for all $t$.
The mathematical formulation for observational data as in \eqref{eq:data} is commonly used \citep{CressieWikle15} and does not preclude uniqueness of inferred parameters despite realizations of noise being nondeterministic.
Note that the second variable $v$ of the ODE \eqref{eq:fhn} is excluded from the data because
typically only the membrane potential can be observed in experiments.
In a statistical/Bayesian framework \citep{KaipioSomersalo05}, the inverse problem translates to finding the posterior probability density $\pi(\boldsymbol{\theta} \,\vert\, \boldsymbol{d})$ of the parameters $\boldsymbol{\theta}$ given data $\boldsymbol{d}$, where $\boldsymbol{d}$ is a discretization of $d(t)$. %
The posterior is, via Bayes' rule, composed of a likelihood term $\pi(\boldsymbol{d} \,\vert\, \boldsymbol{\theta})$ and a prior term $\pi(\boldsymbol{\theta})$,
\begin{equation}
  \label{eq:bayes}
  \pi(\boldsymbol{\theta} \,\vert\, \boldsymbol{d}) \propto \pi(\boldsymbol{d} \,\vert\, \boldsymbol{\theta}) \, \pi(\boldsymbol{\theta}).
\end{equation}

In this work, we assume little knowledge about the range of parameter values $\boldsymbol{\theta}$, hence we target prior distributions that are merely weakly informative (i.e., relatively large variance of $\pi(\boldsymbol{\theta})$).
In order to construct a weak prior, we consider that parameters of the FitzHugh--Nagumo model are in practice chosen to be inside the interval $\theta_0,\theta_1 \in [0,1]$.  Note that the parameter bounds are not required for solvability of the FitzHugh--Nagumo equations \eqref{eq:fhn}.
Due to these bounds, our prior specifies that each parameter is normal and i.i.d.\ (independent identically distributed),
\begin{equation}
  \label{eq:prior}
  \theta_0 \sim
    \mathcal{N}\left(\bar{\theta}_{0,\text{prior}}, \sigma_{0,\text{prior}}^2\right) =
    \mathcal{N}\left(0.4, 0.3^2\right),
  \quad%
  \theta_1 \sim
    \mathcal{N}\left(\bar{\theta}_{1,\text{prior}}, \sigma_{1,\text{prior}}^2\right) =
    \mathcal{N}\left(0.4, 0.4^2\right),
\end{equation}
where
$[0,1] \subset [\bar{\theta}_{i,\text{prior}}-2\sigma_{i,\text{prior}}, \bar{\theta}_{i,\text{prior}}+2\sigma_{i,\text{prior}}]$, $i=0,1$.
Additionally, we limit the range of the parameters to be inside the intervals
\begin{equation}
  \label{eq:bounds}
  \theta_0 \in [-0.2, 1.0] \quad\text{and}\quad
  \theta_1 \in [-0.4, 1.2]
\end{equation}
by means of rejecting prior samples outside of these bounds.
The prior bounds \eqref{eq:bounds} restrict samples of $\boldsymbol{\theta}$ to ranges that contain the unit interval, and, moreover, they limit combinations of parameters where the FitzHugh--Nagumo model generates zero spikes or just one spike (see Appendix~\ref{sec:parameters}).

For the NN-based solution techniques %
proposed in the following Section~\ref{sec:networks}, we target point estimates of
$\boldsymbol{\theta}$ directly.
We observe an analogy in the context of Bayesian inference,
since one type of point estimate of the posterior density is the maximum a posteriori (MAP) point \citep{CalvettiSomersalo07, Stuart10}.  The MAP point is obtained by minimizing the negative log of \eqref{eq:bayes},
\begin{equation}
  \label{eq:map}
  \hat{\boldsymbol{\theta}}_{\textsc{MAP}}
  =
    \argmin_{\boldsymbol{\theta}} \Bigl[-\log \bigl( \pi(\boldsymbol{d} \,\vert\, \boldsymbol{\theta}) \, \pi(\boldsymbol{\theta}) \bigr)\Bigr]
  =
    \argmin_{\boldsymbol{\theta}} \tfrac12 \left\|\frac{d(t) - u_{\boldsymbol{\theta}}(t)}{\sigma_\text{noise}}\right\|_{L_2}^2 \hspace{-1ex} + \tfrac12 \left|\boldsymbol{\theta} - \boldsymbol{\bar\theta}_\text{prior}\right|_{\Sigma_\text{prior}^{-2}}^2,
\end{equation}
where $\sigma_\text{noise}$ denotes the standard deviation of the data noise, which is related to $\eta$ in \eqref{eq:data};
$\boldsymbol{\bar\theta}_\text{prior} \coloneqq (\bar{\theta}_{0,\text{prior}}, \bar{\theta}_{1,\text{prior}})$ is the prior mean, and
$\Sigma_\text{prior} \coloneqq \diag(\sigma_{0,\text{prior}}, \sigma_{1,\text{prior}})$ is the prior standard deviation.
The resulting highly nonlinear loss function pertaining to minimization \eqref{eq:map} is shown in Figure~\ref{fig:ode-loss} (right), %
presenting significant computational challenges, in particular for gradient-based solvers.
Note that our proposed NN-based reconstruction maps are not guaranteed to compute the MAP estimate or other known estimates, such as, maximum likelihood and conditional mean.  The discussion of the MAP estimate serves as an illustration of computational challenges and as an analogy to interpret the results provided by NNs in  Section~\ref{sec:networks}.

\subsection{Deep neural networks for inverse problems}
\label{sec:networks}

We construct NNs to computationally learn reconstruction maps of parameter estimation problems.  We set up the training data such
that the inputs of the NN are time series \eqref{eq:data} coming from the
solution of the ODE \eqref{eq:fhn} and the outputs are the corresponding
parameters $\boldsymbol{\theta}$ that generated the time series.  Since we train the NN to
fit a mapping of time series to parameters in the reverse order of the
forward problem, the NN is learning to represent a pseudoinverse of the
forward operator \citep{AdlerOktem17, FanBohorquezYing19, KhooYing19}, hence approximately solving an inverse problem.
We define our NN-based reconstruction map as
\begin{equation}
  \label{eq:nn-reconstr}
  \hat{\boldsymbol{\theta}} \coloneqq y_L, \quad
  y_\ell = \mathcal{F}_\ell(y_{\ell-1}) \text{ for } 1 \le \ell \le L, \quad
  y_0 \coloneqq \boldsymbol{d},
\end{equation}
where the NN input $\boldsymbol{d}$ is a discrete version of the time series \eqref{eq:data}, $\mathcal{F}_\ell$ denotes layer $\ell$ of the NN, and $\hat{\boldsymbol{\theta}}$ is the network's output, which are predicted model parameters.

\paragraph{Training and testing data for neural networks}

We generate training and testing data for NNs by sampling parameters $\boldsymbol{\theta}$
from their prior distributions \eqref{eq:prior} and discarding samples outside of prior bounds
\eqref{eq:bounds}.  By choosing samples from the prior to generate training
data, we want the NN reconstruction \eqref{eq:nn-reconstr} to learn the most important features of our inverse problem according to prior knowledge.
Furthermore, using the prior for generating training data creates a framework that is analogous to
the original Bayesian inverse problem \eqref{eq:bayes}.
Having obtained samples of $\boldsymbol{\theta}$, we solve the ODE \eqref{eq:fhn} and store the membrane potential $u_{\boldsymbol{\theta}}(t_i)$ at prescribed time steps, $t_i$, $i=1,\ldots,N_t$, with uniform step size $\Delta_t$.
The next two paragraphs describe the different architectures for the layers denoted by $\mathcal{F}_\ell$ in Equation~\eqref{eq:nn-reconstr}.

\paragraph{Dense neural network (dense NN)}

Our first NN architectures consist of a sequence of dense, or fully connected, layers \citep{Goodfellow2016}.  Each dense layer $\ell$ consists of $n_u=n_u(\ell)$ units, or nodes, and each unit of one layer is connected to all other units of a neighboring layer. One dense layer is composed of an affine mapping and a nonlinear function,
$ %
  \mathcal{F}_\ell(y_{\ell-1}) = \phi(W_\ell\,y_{\ell-1} + b_\ell),
$ %
where the matrix
  $W_\ell\in\mathrm{R}^{n_u(\ell) \times n_u(\ell-1)}$
are the weights and
  $b_\ell\in\mathrm{R}^{n_u(\ell)}$
is the bias of layer $\ell$.  The nonlinear activation function $\phi$ is applied element-wise.  A typical choice for $\phi$ is the rectified linear unit (ReLU) function \citep{schmidhuber2015}.  We, however, utilize a smoother activation similar to ReLU, called Swish \citep{ElfwingUchibeDoya18}.  NNs with Swish activation have been shown to suffer less from vanishing gradient problems compared with ReLU \citep{HayouDoucetRousseau2018}, and this behavior has been observed by the authors when training deeper networks.
For our numerical experiments in Section~\ref{sec:results}, we use dense NNs with 2--16 layers and 4--128 units $n_u$ in all layers
to cover a wide range of network sizes, which we will demonstrate to be sufficient for accurate approximation of reconstruction maps.
The dense NNs amount to a total of between 4,034 and 442,114 trainable NN parameters in the form of weight matrices $W_\ell$ and biases $b_\ell$, which are optimized with stochastic gradient descent by using the Adam algorithm \citep{KingmaBa15} and the mean squared error (MSE) loss function (see Section~\ref{sec:results-setup} for more details).

\paragraph{Convolutional neural network (CNN)}

CNNs have been successfully used in two-dimensional image-processing tasks
\citep{lecun1999}.
We take advantage of
the local dependence structure inherent in CNNs to fit
reconstruction maps that have one-dimensional time series with uniform time steps as input.  We therefore exploit locality in our time series with convolutional layers.
A convolutional layer is composed of $n_f$ filters, and each filter is associated with one kernel that is applied to small sections of the time series.  The $n_f$ kernels of one convolutional layer are connected to all neighboring layers, similarly to the units of dense layers described above.  The weights of these connections constitute the NN parameters to be optimized.
Because the weights are shared across time,
significantly fewer weights
have to be optimized.

Our convolutional architectures borrow from CNNs for image classification \citep{KrizhevskySutskeverHinton12}, where we interleave convolutional layers with pooling layers (also called downsampling), which aggregate a small block from a convolutional step into a single value.
Each convolutional layer,
  $y_\ell=\mathcal{F}_\ell(y_{\ell-1})$,
reduces the size of output vectors $y_\ell$ compared with input $y_{\ell-1}$, here by a factor of 1/2; the same reduction holds for pooling layers.  The reduction in size is complemented with an increase of $n_f$ filter counts, here by a factor of 2, from one convolutional layer to the next.
After a few combinations of convolution and pooling, the output is passed to a sequence of dense layers, which concludes the CNN.
Our CNNs are equipped with 2--4 pairs of convolution and pooling layers with varying numbers of filters (see Section~\ref{sec:results-exploration} for more details), and we keep the dense layers at the end of the network fixed to two layers with $n_u=32$ units.  These CNNs achieve NN parameter counts between 5,278 and 261,442.  We utilize the same Swish activation function and optimization setup as for dense NNs described above.

\subsection{Evaluation metrics for inference results}
\label{sec:metrics}

To assess the quality of the prediction from NNs, we carry out both qualitative and quantitative evaluations in Section \ref{sec:results}.
Qualitative assessments are done through scatterplots and time series plots, and the quantification of prediction capabilities is evaluated with several metrics.

In particular, we decompose the commonly used mean squared error (MSE) into its squared bias and centered MSE (C-MSE) components to assess the respective contributions of the mean and of the fluctuations (variability) of the prediction mismatch \citep{taylor2001}:
\vspace{-2ex}
$$
  \MSE(\boldsymbol{\theta},\hat{\boldsymbol{\theta}}) \coloneqq
    \frac{1}{M}\sum_{j=1}^{M} \left(\boldsymbol{\theta}^{(j)} - \hat{\boldsymbol{\theta}}^{(j)}\right)^{2} =
    \overbrace{
      \left(\bar{\boldsymbol{\theta}}-\bar{\hat{\boldsymbol{\theta}}}\right)^2
    }^{=\text{Squared bias}} +
    \overbrace{
      \frac{1}{M}\sum_{j=1}^{M} \left[\Bigl(\boldsymbol{\theta}^{(j)} - \bar{\boldsymbol{\theta}}\Bigr) - \Bigl(\hat{\boldsymbol{\theta}}^{(j)} - \bar {\hat{\boldsymbol{\theta}}}\Bigr)\right]^{2}
    }^{=\text{C-MSE}},
\vspace{-2ex}
$$
where $\boldsymbol{\theta}$ is the true model parameter, $\hat{\boldsymbol{\theta}}$ is the predicted parameter, $\bar{\boldsymbol{\theta}}$ (resp. $\bar{\hat{\boldsymbol{\theta}}}$) is the mean of true parameters (resp. estimated parameters),
and $M$ is the training set size.
We use superscript notation (e.g., $\boldsymbol{\theta}^{(j)}$) to identify different parameters in the training set.
A smaller MSE implies lower errors and hence better predictions. %

Since MSE has the disadvantage of being sensitive to large errors, we report the median of the absolute percentage error (Median-APE),
  \smash{$\bigl|\boldsymbol{\theta}^{(j)}-\hat{\boldsymbol{\theta}}^{(j)}\bigr| / \bigl|\boldsymbol{\theta}^{(j)}\bigr|$}.
This also provides a scale-invariant metric that is less sensitive to outliers than a mean statistic is.
The coefficient of determination $\Rsq$ is used to assess the percentage of variability explained by the prediction
$
 \Rsq \coloneqq 1 -
   \left.  %
     {\sum _{j=1}^{M} \bigl(\boldsymbol{\theta}^{(j)}-\hat{\boldsymbol{\theta}}^{(j)}\bigr)^{2}}
   \middle/  %
     {\sum _{j=1}^{M} \bigl(\boldsymbol{\theta}^{(j)}-{\bar {\boldsymbol{\theta}}}\bigr)^{2}}
    \right. .  %
$
The $\Rsq$ takes values between $-\infty$ and $1$; the closer \Rsq\ is to the ideal value $1$, the more variability is explained by the prediction.

\section{Numerical experiments}\label{sec:results}

In this section we numerically analyze the accuracy of model parameter estimates from dense NNs and CNNs defined in Section~\ref{sec:networks}.  We investigate the effects of different network architectures on the prediction errors of FitzHugh--Nagumo model parameters from noise-free time series data, and we analyze the sensitivity of predictions in the presence of noise in training and/or testing data.  For these numerical studies, we utilize the metrics described in Section~\ref{sec:metrics}.

\subsection{Experimental setup}
\label{sec:results-setup}

The implementation of our algorithms is carried out in Python.  We use the explicit Runge--Kutta method of order 3(2) (RK23) for time integration of the FitzHugh--Nagumo ODE, and we utilize TensorFlow/Keras (version 2.3.1) for implementing the neural networks.
To generate data sets for training, validating, and testing the networks, we simulate the ODE \eqref{eq:fhn} for 200 milliseconds and store the membrane potential at equidistant steps every $\Delta_t=0.2$, which results in a time series of size $N_t=1000$ for each model parameter $\boldsymbol{\theta}\in\mathrm{R}^2$.
Additionally, we use the Fourier transform of the time series for training and prediction in Sections~\ref{sec:results-window} and~\ref{sec:results-noise-prediction}.
We randomly generate sets of training, validating, and testing data based on the prior of $\boldsymbol{\theta}$ as described in Section~\ref{sec:methods}.
Training is performed with $N=1000$ samples unless otherwise specified.  $2000$ samples are used for validation and testing is carried out with a sample size of $M=2000$.
Table~\ref{tab:alg-settings} provides a summary of the experimental settings regarding the optimization, and Table~\ref{tab:nn-settings} shows settings for network architectures.

\begin{table}
  \begin{minipage}[t]{0.44\columnwidth}\centering
  \caption{Settings for training/testing data and optimization algorithms.}
  \label{tab:alg-settings}
  \footnotesize
  \begin{tabular}[t]{lc}
    \toprule
    \thead{Training \& optimization setting} & \thead{Value} \\
    \midrule
    Number of training samples ($N$)      & 1000 \\
    Number of testing samples ($M$)       & 2000 \\
    Number of validating samples          & 2000 \\
    Size of time series ($N_t$)           & 1000 \\
    \midrule
    Loss function                         & MSE \\
    Optimizer                             & Adam \\
    Learning rate (or step length)        & 0.002 \\
    Batch size                            & 32 \\
    Number of epochs with noise-free data & 200 \\
    Number of epochs with noisy data      & 50 \\
    \bottomrule
  \end{tabular}
  \end{minipage}
  \hfill
  \begin{minipage}[t]{0.52\columnwidth}\centering
  \caption{Settings for dense and convolutional neural network architectures.}
  \label{tab:nn-settings}
  \footnotesize
  \begin{tabular}[t]{lc}
    \toprule
    \thead{Neural network setting} & \thead{Value} \\
    \midrule
    Activation function (dense NN \& CNN) & Swish \\
    \midrule
    CNN kernel size                       & 3 \\
    CNN kernel stride                     & 2 \\
    CNN pooling type                      & average \\
    CNN pooling size                      & 2 \\
    CNN pooling stride                    & 2 \\
    CNN padding                           & none \\
    CNN flattening into dense layers      & \hspace{-2em} 2 layers, $n_u=32$ \\
    \bottomrule
  \end{tabular}
  \end{minipage}
\end{table}

\subsection{Exploration of neural network architectures}
\label{sec:results-exploration}

In the following, we investigate the influence of different NN architectures
from Section~\ref{sec:networks} on predictions performed on the validation data set.
We explore effects on predictive skills while varying the number of dense layers and number of units per layer, in the case of dense NNs, and the number of convolutional layers and filters per layer, in the case of CNNs.
The training of the networks as well as predictions are performed on
noise-free time series data, which represents an ideal scenario for parameter estimation.  Noise-free data have
the advantage that optimization algorithms, using stochastic gradient descent, can run for sufficiently high iteration counts (or epochs),
without leading to overfitting due to noise in the data.  This allows us to train
the networks relatively well and to investigate differences in
predictions that are largely due to network architectures.  Prediction results
with noisy observational data are crucial in practice and are
presented in Section~\ref{sec:results-noise}.

The results in this section are augmented with cross-validation experiments and a study on the sensitivity on the initialization of NN parameters for selected NN architectures in Appendix~\ref{sec:crossvalidation}.

\begin{table}
  \caption{Squared bias (C-MSE) of model parameter predictions %
    (noise-free), using \textbf{dense NNs} with variable numbers of layers and units per layer $n_u$.
    }
  \label{tab:denseNN-N1000-cmse}
  \centering
  %
% dense NN
% N=1000
% Sq-Bias (C-MSE)
%
\def\ccR{\cellcolor{jr@red!40}}
\def\ccG{\cellcolor{jr@green!40}}
\footnotesize
\setlength{\tabcolsep}{0.4em}  % set horizontal cell padding
\begin{tabular}{cccccc}
  \toprule
  % table header
  \thead{$n_u$} & \thead{2 layers} & \thead{4 layers} & \thead{8 layers} & \thead{12 layers} & \thead{16 layers} \\
  \midrule
  \thead{4}   &\ccR\spp{0.00012255} (\fppp{0.01538955})&\spp{0.00001018} (\fpp{0.00851010}) &\ccR\spp{0.00035975} (\fppp{0.02089234})&\ccR\spp{0.00049520} (\fppp{0.03039774})&\ccR\spp{0.00010354} (\fppp{0.02687869})\\
  \thead{8}   &\spp{0.00012296} (\fpp{0.00314618}) &\spp{0.00010000} (\fpp{0.00592315}) &\spp{0.00028274} (\fpp{0.00237435}) &\spp{0.00003241} (\fpp{0.00360076}) &\ccR\spp{0.00015784} (\fppp{0.02253452})\\
  \thead{16}  &\ccG\spp{0.00005237} (\fpp{0.00233358}) &\ccG\spp{0.00006155} (\fpp{0.00232216}) &\ccG\spp{0.00008577} (\fpp{0.00167592}) &\spp{0.00001555} (\fpp{0.00304074}) &\spp{0.00013031} (\fpp{0.00331612}) \\
  \thead{32}  &\ccG\spp{0.00000843} (\fpp{0.00202809}) &\ccG\spp{0.00003737} (\fpp{0.00204163}) &\ccG\spp{0.00001084} (\fpp{0.00166543}) &\ccG\spp{0.00002532} (\fpp{0.00189084}) &\spp{0.00002568} (\fpp{0.00252210}) \\
  \thead{64}  &\spp{0.00010551} (\fpp{0.00264516}) &\ccG\spp{0.00006433} (\fpp{0.00215075}) &\ccG\spp{0.00009752} (\fpp{0.00184323}) &\spp{0.00020403} (\fpp{0.00325581}) &\spp{0.00000859} (\fpp{0.00255593}) \\
  \thead{128} &\spp{0.00052733} (\fpp{0.00895563}) &\ccG\spp{0.00000970} (\fpp{0.00147920}) &\ccG\spp{0.00004914} (\fpp{0.00223369}) &\spp{0.00002290} (\fpp{0.00264401}) &\spp{0.00010787} (\fpp{0.00227690}) \\
  \bottomrule
\end{tabular}

  \caption{Median-APE (\Rsq) of model parameter predictions %
    (noise-free), using \textbf{dense NNs}.
    }
  \label{tab:denseNN-N1000-r2}
  \centering
  \vspace{0.5ex}%
  %
% dense NN
% N=1000
% Med-APE (R^2)
%
\def\ccR{\cellcolor{jr@red!40}}
\def\ccG{\cellcolor{jr@green!40}}
\footnotesize
\begin{tabular}{cccccc}
  \toprule
  % table header
  \thead{$n_u$} & \thead{2 layers} & \thead{4 layers} & \thead{8 layers} & \thead{12 layers} & \thead{16 layers} \\
  \midrule
  \thead{4}   &\ccR\fppp{0.17597809} (\fppp{0.83459631}) &\fpp{0.06519552} (\fppp{0.91583440}) &\ccR\fppp{0.10281777} (\fppp{0.79302538}) &\ccR\fppp{0.23903932} (\fppp{0.69325031}) &\ccR\fppp{0.20294505} (\fppp{0.73237926}) \\
  \thead{8}   & \fpp{0.05888217} (\fppp{0.96813391}) &\fpp{0.05086597} (\fppp{0.94370045}) & \fpp{0.04892407} (\fppp{0.97256795}) & \fpp{0.04202925} (\fppp{0.96548110}) &\ccR\fppp{0.18166484} (\fppp{0.78678876}) \\
  \thead{16}  &\ccG\fpp{0.03974951} (\fppp{0.97588337}) &\ccG\fpp{0.04434419} (\fppp{0.97530346}) &\ccG\fpp{0.03188437} (\fppp{0.98198332}) & \fpp{0.02904615} (\fppp{0.97086141}) & \fpp{0.04621948} (\fppp{0.96665952}) \\
  \thead{32}  &\ccG\fpp{0.02931040} (\fppp{0.97909389}) &\ccG\fpp{0.02546401} (\fppp{0.97876271}) &\ccG\fpp{0.02623869} (\fppp{0.98265849}) &\ccG\fpp{0.02601104} (\fppp{0.98046204}) & \fpp{0.05868318} (\fppp{0.97437841}) \\
  \thead{64}  & \fpp{0.04291739} (\fppp{0.97226239}) &\ccG\fpp{0.02905429} (\fppp{0.97720732}) &\ccG\fpp{0.03796276} (\fppp{0.97968331}) & \fpp{0.06253323} (\fppp{0.96651542}) & \fpp{0.04518705} (\fppp{0.97414724}) \\
  \thead{128} & \fpp{0.08187380} (\fppp{0.90542436}) &\ccG\fpp{0.01936326} (\fppp{0.98485333}) &\ccG\fpp{0.03529748} (\fppp{0.97650937}) & \fpp{0.05123023} (\fppp{0.97300799}) & \fpp{0.05217933} (\fppp{0.97623276}) \\
  \bottomrule
\end{tabular}

\end{table}

\paragraph{Model parameter predictions with dense NNs}

The prediction capability of dense NNs are evaluated with the squared bias
and C-MSE metrics in Table~\ref{tab:denseNN-N1000-cmse} for the number of dense
layers varying from 2 to 16 and the number of units per layer
$n_u$ between 4 and 128.  Corresponding evaluations with the Median-APE and \Rsq\ metrics are
shown in Table~\ref{tab:denseNN-N1000-r2}.  The tables highlight in red those configurations
leading to relatively poor predictions and in green those leading to relatively good
predictions.
Overall, we observe a low error in average squared bias (below $10^{-3}$) and
mild variations of the errors, expressed in C-MSE values below $10^{-2}$, showing that a large contribution to the total MSE is due to the mismatch in variability between true and predicted parameters.
Additionally, the scale invariant metric Median-APE assumes values mostly below
$10^{-1}$ and the \Rsq\ metric mostly above $0.95$.
From these tables, we can deduce that networks with 2--12 layers and 16--64
units per layer lead to the lowest prediction errors (green cells in
Tables~\ref{tab:denseNN-N1000-cmse}, \ref{tab:denseNN-N1000-r2}).  Networks
with small unit counts, such as 4, however, deteriorate in their
prediction skills, because they do not offer sufficient degrees of freedom to represent a reconstruction map.
Furthermore, deeper networks with large numbers of layers, such as 16, do not
offer any improvements of prediction errors compared with shallower architectures.
Due to these results, we select the dense NN architecture with 4 layers and $n_u=32$ units per layer for cross-validation and inspection of sensitivity to initialization of NN parameters (see Appendix~\ref{sec:crossvalidation}); and we carry out subsequent experiments with this choice of dense NN.
Next, we improve upon the already good results of dense NNs using
convolutional layers.

\begin{table}
  \caption{Squared bias (C-MSE) of model parameter predictions %
    (noise-free),
    using \textbf{CNNs} with variable numbers of convolutional layers and filters per
    layer $n_f$. %
    }
  \label{tab:CNN-N1000-cmse}
  \centering
  %
% CNN
% N=1000
% Sq-Bias (C-MSE)
%
\def\ccR{\cellcolor{jr@red!40}}
\def\ccG{\cellcolor{jr@green!40}}
\footnotesize
\begin{tabular}{cccc}
  \toprule
  % table header
  \thead{$n_f$} & \thead{$n_f\times[1, 2]$} & \thead{$n_f\times[1, 2, 4]$} & \thead{$n_f\times[1, 2, 4, 8]$} \\
  \midrule
  % table body
% results 2020-11-22
% \thead{2}  & \spp{0.00006317} (\fpp{0.00066688}) & \spp{0.00007876} (\fpp{0.00055230}) & \ccG\spp{0.00003243} (\fpp{0.00050343}) \\
% \thead{4}  & \ccR\spp{0.00001270} (\fppp{0.00160141}) & \ccG\spp{0.00006540} (\fpp{0.00051545}) & \ccG\spp{0.00006438} (\fpp{0.00040737}) \\
% \thead{8}  & \ccR\spp{0.00014845} (\fppp{0.00128673}) & \ccG\spp{0.00001886} (\fpp{0.00038365}) & \ccG\spp{0.00003297} (\fpp{0.00041710}) \\
% \thead{16} & \ccR\spp{0.00045813} (\fppp{0.00134662}) & \spp{0.00001228} (\fpp{0.00065489}) & \spp{0.00002379} (\fpp{0.00060042}) \\
% \thead{32} & \ccR\spp{0.00029676} (\fppp{0.00103814}) & \spp{0.00000362} (\fpp{0.00057423}) & \spp{0.00000585} (\fpp{0.00058385}) \\
% results 2021-02-23b
  \thead{2}  &\spp{0.00005219} (\fpp{0.00089193}) &\spp{0.00001278} (\fpp{0.00061011}) &\ccG\spp{0.00000066} (\fpp{0.00056402}) \\
  \thead{4}  &\spp{0.00004528} (\fpp{0.00069384}) &\ccG\spp{0.00000422} (\fpp{0.00047806}) &\ccG\spp{0.00002749} (\fpp{0.00044766}) \\
  \thead{8}  &\spp{0.00000030} (\fpp{0.00070707}) &\ccG\spp{0.00001848} (\fpp{0.00048323}) &\ccG\spp{0.00002030} (\fpp{0.00044412}) \\
  \thead{16} &\spp{0.00002182} (\fpp{0.00066050}) &\spp{0.00004453} (\fpp{0.00054422}) &\spp{0.00003001} (\fpp{0.00065078}) \\
  \thead{32} &\spp{0.00001654} (\fppp{0.00107728})&\spp{0.00025069} (\fpp{0.00063772}) &\spp{0.00013898} (\fpp{0.00050925}) \\
  \bottomrule
\end{tabular}

  \caption{Median-APE (\Rsq) of model parameter predictions %
    (noise-free),
    using \textbf{CNNs}.
  }
  \label{tab:CNN-N1000-r2}
  \centering
  \vspace{0.5ex}%
  %
% CNN
% N=1000
% Med-APE (R^2)
%
\def\ccR{\cellcolor{jr@red!40}}
\def\ccG{\cellcolor{jr@green!40}}
\footnotesize
\begin{tabular}{cccc}
  \toprule
  % table header
  \thead{$n_f$} & \thead{$n_f\times[1, 2]$} & \thead{$n_f\times[1, 2, 4]$} & \thead{$n_f\times[1, 2, 4, 8]$} \\
  \midrule
  % table body
% results 2020-11-22
% \thead{2}  & \fpp{0.02505420} (\fppp{0.99232556}) & \fpp{0.02854986} (\fppp{0.99317710}) & \ccG\fpp{0.01662370} (\fppp{0.99415575}) \\
% \thead{4}  & \ccR\fpp{0.02107035} (\fppp{0.98430298}) & \ccG\fpp{0.01844387} (\fppp{0.99393668}) & \ccG\fpp{0.02442103} (\fppp{0.99494346}) \\
% \thead{8}  & \ccR\fpp{0.03194577} (\fppp{0.98493964}) & \ccG\fpp{0.01678852} (\fppp{0.99543745}) & \ccG\fpp{0.02114569} (\fppp{0.99481827}) \\
% \thead{16} & \ccR\fpp{0.04154356} (\fppp{0.98174011}) & \fpp{0.02137104} (\fppp{0.99277213}) & \fpp{0.03058852} (\fppp{0.99321893}) \\
% \thead{32} & \ccR\fpp{0.03164306} (\fppp{0.98691046}) & \fpp{0.02198170} (\fppp{0.99336539}) & \fpp{0.02193088} (\fppp{0.99311787}) \\
% results 2021-02-23b
  \thead{2}  &\fpp{0.02229769} (\fppp{0.98995082}) &\fpp{0.01690022} (\fppp{0.99303543}) &\ccG\fpp{0.01682202} (\fppp{0.99400382}) \\
  \thead{4}  &\fpp{0.02276215} (\fppp{0.99225350}) &\ccG\fpp{0.01604877} (\fppp{0.99470133}) &\ccG\fpp{0.01340548} (\fppp{0.99489327}) \\
  \thead{8}  &\fpp{0.01746480} (\fppp{0.99241229}) &\ccG\fpp{0.01842535} (\fppp{0.99433798}) &\ccG\fpp{0.01593072} (\fppp{0.99504607}) \\
  \thead{16} &\fpp{0.01783907} (\fppp{0.99262075}) &\fpp{0.02235540} (\fppp{0.99345349}) &\fpp{0.02617287} (\fppp{0.99314100}) \\
  \thead{32} &\fpp{0.02107835} (\fppp{0.98833119}) &\fpp{0.03225408} (\fppp{0.99118147}) &\fpp{0.03339433} (\fppp{0.99322037}) \\
  \bottomrule
\end{tabular}

\end{table}

\paragraph{Model parameter predictions with CNNs}

Analogous to the results with dense NNs, we now consider
CNNs while changing the number of convolutional layers and the number of
filters $n_f$ per layer.  Two dense layers with 32 units follow the
convolutional layers and remain fixed during these experiments.
Table~\ref{tab:CNN-N1000-cmse} shows evaluations with squared bias and C-MSE
metrics, and Table~\ref{tab:CNN-N1000-r2} considers the Median-APE and \Rsq\
metrics.
For brevity, we express the convolutional layers as a list, for instance
  $n_f\times[1,2,4]$,
which denotes $(n_f\times1)$ filters in the first layer, $(n_f\times2)$
filters in the second layer, and $(n_f\times4)$ filters in the third layer.
Tables~\ref{tab:CNN-N1000-cmse} and~\ref{tab:CNN-N1000-r2} show that CNNs can further reduce the prediction errors for our inverse problem compared with
dense NNs.  This improvement is demonstrated by C-MSE values below $10^{-3}$ (one order of
magnitude lower than with dense NNs) and \Rsq\ values mostly above $0.99$
(previously $0.95$--$0.98$ with dense NNs), which is very close to the ideal
\Rsq\ of 1.0.
These improvements relative to dense NN can be attributed to the
CNNs exploiting local dependence information of neighboring points in the time series.

We additionally observe that CNNs with only two convolutional layers yield
higher prediction errors than with three and four layers.
Thus the short networks do not offer sufficient degrees of freedom to adapt to
the reconstruction map of the inverse problem. %
Moreover, increasing the number of filters to $n_f=16$ and beyond %
shows no benefit for predictive skills.  The best results are
achieved with 3--4 convolutional layers and $n_f=4,\ldots,8$, %
which we highlight with green in the tables.
Because of the results, we select the CNN consisting of 3 convolutional layers with 8, 16, and 32 filters, in short denoted by $n_f\times[1,2,4]$, $n_f=8$, for cross-validation and inspection of sensitivity to initialization of NN parameters (see Appendix~\ref{sec:crossvalidation}); and we carry out subsequent numerical experiments with this choice of CNN.

\subsection{Learning capabilities of neural networks for partially observed time series}
\label{sec:results-window}

This section demonstrates numerically how the networks are learning as we try to find answers to the higher-level question: Is a NN merely ``remembering'' a time series or is it learning underlying properties or dynamics?
To this end, we design an experiment where the time series of the training data from Section~\ref{sec:results-setup} is split into two halves of length $N_t/2=500$, hence doubling the number of training samples from $N=1000$ to $2000$.
After training, the predictions are performed on time series of the testing data that are also of length $N_t/2=500$.
However, we extract the following five arbitrary overlapping intervals from the original testing data of length $N_t=1000$ (see Section~\ref{sec:results-setup}):
  $[30,530)$,
  $[146,646)$,
  $[174,674)$,
  $[362,862)$,
  $[370,870)$;
thus increasing the amount of testing samples from $M=2000$ to $\num{10000}$.
In addition to using time series data, we use a second input data type by replacing each time series by its Fourier transform; and a third input data type by combining time series and Fourier transform.
The results in Tables~\ref{tab:denseNN-window} and~\ref{tab:CNN-window} clearly show the advantage of CNNs over dense NNs.
While the CNN's accuracy in this setting is similar to the one presented in Section~\ref{sec:results-exploration},
the dense NN performs poorly ($\Rsq<0.5$ with time series data).  The dense network's prediction capability improves when Fourier data is used; however, the model parameter predictions from a Fourier spectrum are significantly worse compared with previous results using (full) time series data (Tables \ref{tab:denseNN-N1000-cmse}, \ref{tab:denseNN-N1000-r2}).
Overall, we conclude from these results that only CNNs demonstrate a capability to extract crucial properties of the time series that are important for inferring model parameters from arbitrary, partial observations.  This is presumably achieved by the shift-invariant kernel of convolutional layers.
Thus, the results suggest that CNNs not simply memorize patterns, but rather recognize properties or dynamics of the ODE.

\begin{table}
  \begin{minipage}{0.48\columnwidth}\centering
  \caption{%
    Predictions with partially observed time series (noise-free),
    using a \textbf{dense NN} with $4$ layers, $n_u=32$;
    performance is poor compared with CNNs.}
  \label{tab:denseNN-window}
  \centering
  \vspace{1ex} %
  %
% CNN
%
\footnotesize
\setlength{\tabcolsep}{0.2em}  % set horizontal cell padding
\begin{tabular}{lcccc}
  \toprule
  % table header
  \thead{Data type} &
  \thead{Sq.\ bias} &
  \thead{C-MSE}     &
  \thead{Med.-APE}  &
  \thead{\Rsq}      \\
  \midrule
  % table body
  Time            & \sppp{0.00351007} & \fppp{0.05336661} & \fpppp{0.26324250} & \fppp{0.47533343} \\
  Fourier         & \sppp{0.00024752} & \fppp{0.03118081} & \fpppp{0.14775001} & \fppp{0.62040182} \\
  Time \& Fourier & \sppp{0.00426909} & \fppp{0.04683503} & \fpppp{0.25025188} & \fppp{0.52793477} \\
  \bottomrule
\end{tabular}

% # N1000_windowOffsetType-default_fftApply0_stateReplace0
% Sq-Bias (C-MSE): 0.00351007 (0.05336661)
% Med-APE (R^2):   0.26324250 (0.47533343)
% # N1000_windowOffsetType-default_fftApply1_stateReplace1
% Sq-Bias (C-MSE): 0.00024752 (0.03118081)
% Med-APE (R^2):   0.14775001 (0.62040182)
% # N1000_windowOffsetType-default_fftApply1_stateReplace0
% Sq-Bias (C-MSE): 0.00426909 (0.04683503)
% Med-APE (R^2):   0.25025188 (0.52793477)
  \end{minipage}
  \hfill
  \begin{minipage}{0.48\columnwidth}\centering
  \caption{%
    Predictions with partially observed time series (noise-free);
    using a \textbf{CNN} with $n_f\times[1,2,4]$, $n_f=8$;
    performance similar as with full time series.}
  \label{tab:CNN-window}
  \centering
  \vspace{1ex} %
  %
% CNN
%
\footnotesize
\setlength{\tabcolsep}{0.2em}  % set horizontal cell padding
\begin{tabular}{lcccc}
  \toprule
  % table header
  \thead{Data type} &
  \thead{Sq.\ bias} &
  \thead{C-MSE}     &
  \thead{Med.-APE}  &
  \thead{\Rsq}      \\
  \midrule
  % table body
  Time            & \sppp{0.00008383} & \fppp{0.00334244} & \fppp{0.02348651} & \fppp{0.96955629} \\
  Fourier         & \sppp{0.00016982} &\fpppp{0.02456009} &\fpppp{0.12346910} & \fppp{0.68519956} \\
  Time \& Fourier & \sppp{0.00000859} & \fppp{0.00221823} & \fppp{0.02887204} & \fppp{0.97953297} \\
  \bottomrule
\end{tabular}

% # N1000_windowOffsetType-default_fftApply0_stateReplace0
% Sq-Bias (C-MSE): 0.00008383 (0.00334244)
% Med-APE (R^2):   0.02348651 (0.96955629)
% # N1000_windowOffsetType-default_fftApply1_stateReplace1
% Sq-Bias (C-MSE): 0.00016982 (0.02456009)
% Med-APE (R^2):   0.12346910 (0.68519956)
% # N1000_windowOffsetType-default_fftApply1_stateReplace0
% Sq-Bias (C-MSE): 0.00000859 (0.00221823)
% Med-APE (R^2):   0.02887204 (0.97953297)
  \end{minipage}
\end{table}

\subsection{Parameter estimation in the presence of noise in observational data}
\label{sec:results-noise}

\paragraph{Noise model}

We consider an additive noise model that is first-order autoregressive in time and widely used for representing time series.
The autoregressive process is parametrized by a correlation parameter $\rho$, which determines the dependence of the process on its previous value \citep{mills1991time}.
Recall from \eqref{eq:data} that the observational data are
 $d(t) = u_{\boldsymbol{\theta}^\ast}(t) + \eta(t)$,
where $u_{\boldsymbol{\theta}^\ast}(t)$ comes from the solution of \eqref{eq:fhn} and $\eta(t)$ is a correlated noise that evolves in time as
\begin{equation}
  \label{eq:noise-model}
  \eta(t_i) \coloneqq \rho\,\eta(t_{i-1}) + \epsilon(t_i), \quad
  i = 2,\ldots,N_t,
  \quad
  \eta(t_1) \sim \mathcal{N}\Bigl(0,\tfrac{\sigma^2}{\Delta_t^2}\Bigr)
\end{equation}
with $|\rho| < 1$.
The term $\epsilon(t_i)$ is independent from $\eta(t_{i-1})$ and the process $\epsilon$ is a normally distributed white noise.
To reflect the model resolution $\Delta_t$ in the level of variance in noise $\eta$, we prescribe $\var(\eta)=\sigma^2/\Delta_t^2$ together with $\rho$.
The variance of $\eta$ is constant across time since the process is stationary due to $|\rho| < 1$, meaning that the stationary distribution of the process is
  \smash{$\eta(t) \sim \mathcal{N}\bigl(0,\frac{\sigma^2}{\Delta_t^2}\bigr)$}
for all $t$.  Consequently, we derive%
\footnote{
  Note that it is more common to prescribe $\var(\epsilon)$ together with $\rho$ first and then derive $\var(\eta)$; however, we proceed in the reverse order because we are foremost interested in defining the effective variance of $\eta$.}
$\epsilon(t_i) \sim \mathcal{N}\bigl(0,\frac{\sigma^2}{\Delta_t^2} (1-\rho^2)\bigr)$.

In the following, when noisy data are used, the values of the parameter $\rho$ and $\sigma$ are varied randomly across different samples of \eqref{eq:data}.
We generate $100$ independent samples of
\begin{equation}
  \label{eq:noise-prior}
  \rho \sim \mathcal{N}(0.8,0.05^2)
  \quad\text{and}\quad
  \sigma \sim \mathcal{N}(0.07,0.01^2)
\end{equation}
in order to achieve a correlation $\rho$ of $0.65$--$0.95$ and a $\Delta_t$-independent noise level of $4$--$10$\%
For each value of the pair $(\rho,\sigma)$, %
independent replicates of the process $\eta$ following \eqref{eq:noise-model} are generated and added to training and/or testing data.

\begin{table}
  \begin{minipage}{0.48\columnwidth}\centering
  \caption{Median-APE (\Rsq) of model parameter predictions with noisy observational data, using a \textbf{dense NN}
  with $4$ layers, $n_u=32$.}
  \label{tab:denseNN-varyData-r2}
  \vspace{1ex} %
  %
% denseNN
% Med-APE (R^2)
%
\footnotesize
\setlength{\tabcolsep}{0.2em}  % set horizontal cell padding
\begin{tabular}{cccc}
  \toprule
  % table header
  \thead{$N$}
    & \thead{\scalebox{.9}[1.0]{train noise-free}\\\scalebox{.9}[1.0]{test noise-free}}
    & \thead{\scalebox{.9}[1.0]{train noise-free}\\\scalebox{.9}[1.0]{test with noise}}
    & \thead{\scalebox{.9}[1.0]{train with noise}\\\scalebox{.9}[1.0]{test with noise}}

  \\
  \midrule
  % table body
%   \thead{\phantom{1}100}
%     & \fppp{0.18269528} (\fppp{0.70531805})
%     & \fppp{0.23747228} (\fppp{0.62268548})
%     & \fppp{0.29260270} (\fppp{0.51066700})
%     \\
%   \thead{\phantom{1}250}
%     & \fpp{0.07462242} (\fppp{0.90671319})
%     & \fppp{0.12877109} (\fppp{0.86129710})
%     & \fppp{0.15933400} (\fppp{0.78201168})
%     \\
  \thead{\phantom{1}500} & \fpp{0.04254036} (\fppp{0.95992348}) & \fpp{0.09806508} (\fppp{0.91350928}) &\fppp{0.10494704} (\fppp{0.87896072}) \\
  \thead{1000}           & \fpp{0.02102068} (\fppp{0.97766627}) &\fppp{0.10293057} (\fppp{0.91810964}) & \fpp{0.08168079} (\fppp{0.92074530}) \\
  \thead{4000}           & \fpp{0.01439892} (\fppp{0.99308080}) & \fpp{0.08899740} (\fppp{0.92718636}) & \fpp{0.06084910} (\fppp{0.96110537}) \\
  \thead{8000}           & \fpp{0.02121672} (\fppp{0.99221882}) & \fpp{0.09801366} (\fppp{0.92121875}) & \fpp{0.06187165} (\fppp{0.96824691}) \\
  \bottomrule
\end{tabular}

% # N500_noiseTrain0_noiseTest0
% Med-APE (R^2):   0.04254036 (0.95992348)
% # N500_noiseTrain0_noiseTest1
% Med-APE (R^2):   0.09806508 (0.91350928)
% # N500_noiseTrain1_noiseTest1
% Med-APE (R^2):   0.10494704 (0.87896072)
% ========================================
% # N1000_noiseTrain0_noiseTest0
% Med-APE (R^2):   0.02102068 (0.97766627)
% # N1000_noiseTrain0_noiseTest1
% Med-APE (R^2):   0.10293057 (0.91810964)
% # N1000_noiseTrain1_noiseTest1
% Med-APE (R^2):   0.08168079 (0.92074530)
% ========================================
% # N4000_noiseTrain0_noiseTest0
% Med-APE (R^2):   0.01439892 (0.99308080)
% # N4000_noiseTrain0_noiseTest1
% Med-APE (R^2):   0.08899740 (0.92718636)
% # N4000_noiseTrain1_noiseTest1
% Med-APE (R^2):   0.06084910 (0.96110537)
% ========================================
% # N8000_noiseTrain0_noiseTest0
% Med-APE (R^2):   0.02121672 (0.99221882)
% # N8000_noiseTrain0_noiseTest1
% Med-APE (R^2):   0.09801366 (0.92121875)
% # N8000_noiseTrain1_noiseTest1
% Med-APE (R^2):   0.06187165 (0.96824691)

  \end{minipage}
  \hfill
  \begin{minipage}{0.48\columnwidth}\centering
  \caption{Median-APE (\Rsq) of model parameter predictions with noisy observational data, using a \textbf{CNN}
  with $n_f\times[1,2,4]$, $n_f=8$.
  }
  \label{tab:CNN-varyData-r2}
  \vspace{1ex} %
  %
% CNN
% Med-APE (R^2)
%
\footnotesize
\setlength{\tabcolsep}{0.2em}  % set horizontal cell padding
\begin{tabular}{cccc}
  \toprule
  % table header
  \thead{$N$}
    & \thead{\scalebox{.9}[1.0]{train noise-free}\\\scalebox{.9}[1.0]{test noise-free}}
    & \thead{\scalebox{.9}[1.0]{train noise-free}\\\scalebox{.9}[1.0]{test with noise}}
    & \thead{\scalebox{.9}[1.0]{train with noise}\\\scalebox{.9}[1.0]{test with noise}}
  \\
  \midrule
  % table body
  \thead{\phantom{1}500} & \fpp{0.02262855} (\fppp{0.99038866}) & \fppp{0.16948119} (\fppp{0.78793130}) & \fpp{0.09798870} (\fppp{0.92067802}) \\
  \thead{1000}           & \fpp{0.01420267} (\fppp{0.99513618}) & \fppp{0.17401188} (\fppp{0.76276474}) & \fpp{0.09579921} (\fppp{0.93774399}) \\
  \thead{4000}           & \fpp{0.01394978} (\fppp{0.99705447}) & \fppp{0.20429144} (\fppp{0.71005158}) & \fpp{0.05968948} (\fppp{0.96957965}) \\
  \thead{8000}           & \fpp{0.01386153} (\fppp{0.99836268}) & \fppp{0.25123448} (\fppp{0.61708729}) & \fpp{0.05302422} (\fppp{0.97604255}) \\
  \bottomrule
\end{tabular}

% # N500_noiseTrain0_noiseTest0
% Med-APE (R^2):   0.02262855 (0.99038866)
% # N500_noiseTrain0_noiseTest1
% Med-APE (R^2):   0.16948119 (0.78793130)
% # N500_noiseTrain1_noiseTest1
% Med-APE (R^2):   0.09798870 (0.92067802)
% ========================================
% # N1000_noiseTrain0_noiseTest0
% Med-APE (R^2):   0.01420267 (0.99513618)
% # N1000_noiseTrain0_noiseTest1
% Med-APE (R^2):   0.17401188 (0.76276474)
% # N1000_noiseTrain1_noiseTest1
% Med-APE (R^2):   0.09579921 (0.93774399)
% ========================================
% # N4000_noiseTrain0_noiseTest0
% Med-APE (R^2):   0.01394978 (0.99705447)
% # N4000_noiseTrain0_noiseTest1
% Med-APE (R^2):   0.20429144 (0.71005158)
% # N4000_noiseTrain1_noiseTest1
% Med-APE (R^2):   0.05968948 (0.96957965)
% ========================================
% # N8000_noiseTrain0_noiseTest0
% Med-APE (R^2):   0.01386153 (0.99836268)
% # N8000_noiseTrain0_noiseTest1
% Med-APE (R^2):   0.25123448 (0.61708729)
% # N8000_noiseTrain1_noiseTest1
% Med-APE (R^2):   0.05302422 (0.97604255)

  \end{minipage}
\end{table}

\begin{figure}\centering
  \includegraphics[width=0.9\columnwidth, trim=5 20 5 5, clip]{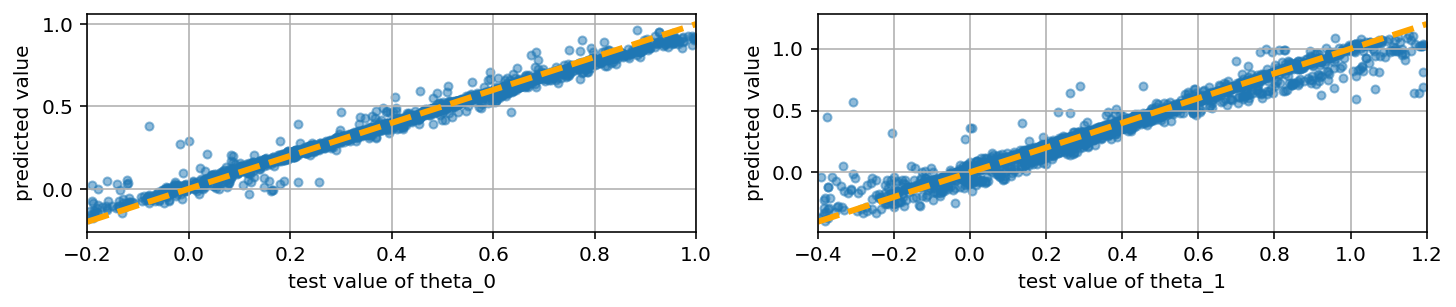}
  \includegraphics[width=0.9\columnwidth, trim=5 20 5 5, clip]{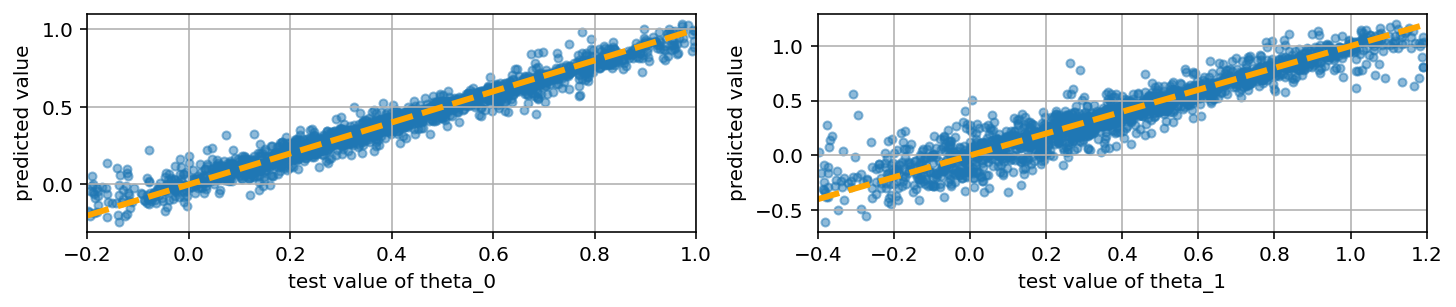}
  \includegraphics[width=0.9\columnwidth, trim=5 0 5 5, clip]{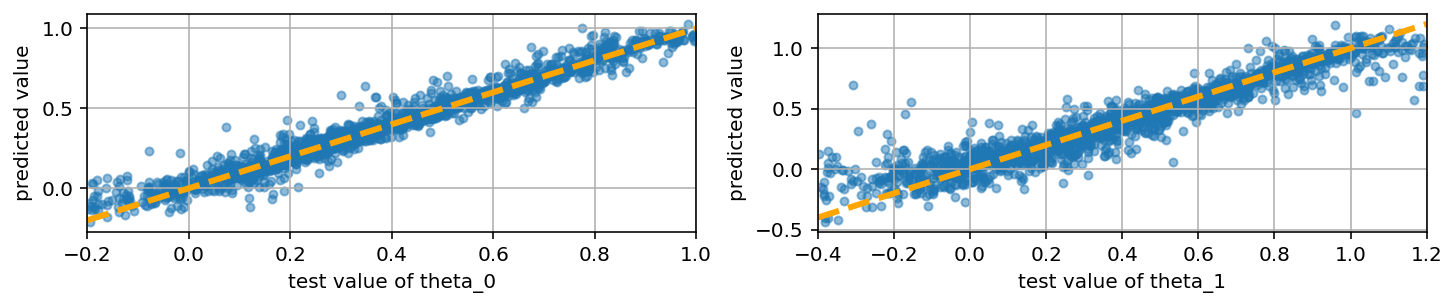}
  \vspace{-2ex}
  \caption{Predictions of parameters $\theta_0$ (left column) and $\theta_1$ (right column) from noise-free and noisy observational data, using a \textbf{dense NN} ($4$ layers, $n_u=32$) and $N=1000$ training samples.
  Top row: training and testing data noise-free;
  Middle row: training data noise-free and testing data with noise;
  Bottom row: training and testing data with noise.
  Dashed orange line depicts perfect predictions.}
  \label{fig:denseNN-varyNoise}
  \vspace{6ex} %
  \includegraphics[width=0.9\columnwidth, trim=5 20 5 5, clip]{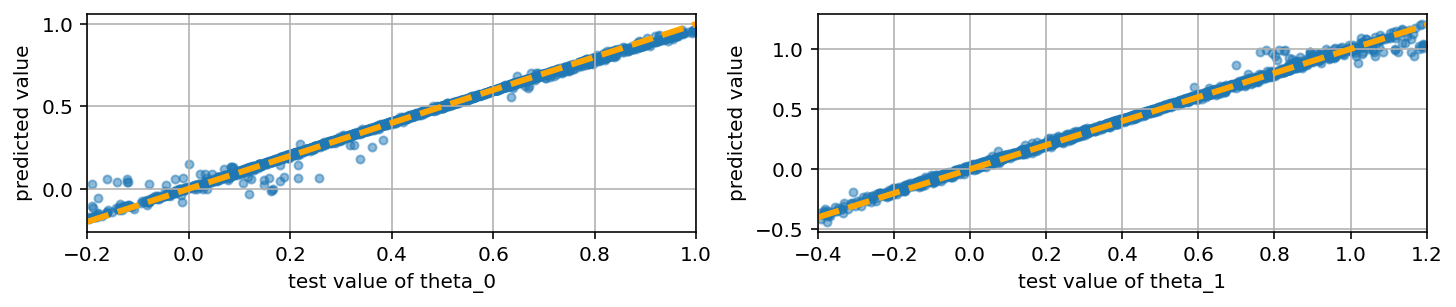}
  \includegraphics[width=0.9\columnwidth, trim=5 20 5 5, clip]{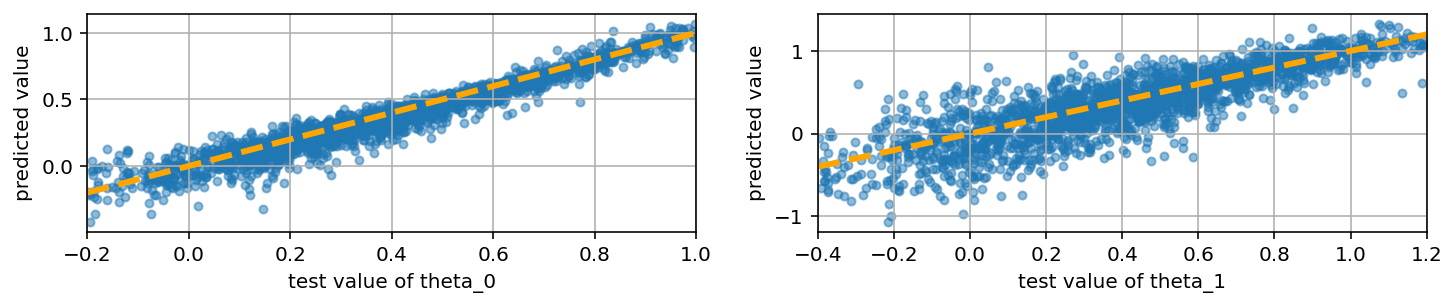}
  \includegraphics[width=0.9\columnwidth, trim=5 0 5 5, clip]{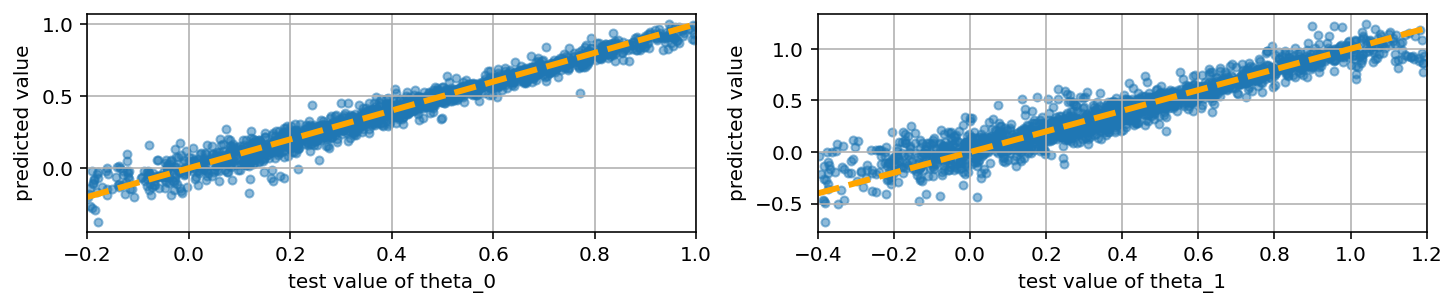}
  \vspace{-2ex}
  \caption{Predictions of parameters $\theta_0$ (left column) and $\theta_1$ (right column) from noise-free and noisy observational data, using a \textbf{CNN} ($n_f\times[1,2,4]$, $n_f=8$) and $N=1000$ training samples.
  Top row: training and testing data noise-free;
  Middle row: training data noise-free and testing data with noise;
  Bottom row: training and testing data with noise.
  Dashed orange line depicts perfect predictions.}
  \label{fig:CNN-varyNoise}
\end{figure}

\paragraph{Model parameter predictions with noise}

More relevant in practice are model parameter estimates in the presence of noise in observational data.
Using one dense NN and one CNN, we demonstrate the effects of the additive noise stemming from the model described above on predictive skills.
For these experiments, we use a dense NN with four layers and $n_u=32$ units per layer, and the CNN has $3$ convolutional layers with $n_f\times[1,2,4]$, $n_f=8$ and $2$ dense layers.
Tables~\ref{tab:denseNN-varyData-r2} and~\ref{tab:CNN-varyData-r2} show the prediction errors for the dense NN and CNN, respectively.  Three columns in each table represent different noise configurations: both training and testing data do not contain noise (second column); only testing data contain noise (third column); and both training and testing data contain noise (fourth column).
The tables additionally demonstrate the effects of using smaller and larger amounts of training data, denoted by $N$ (first column), indicating in most cases improved predictive skills with increasing training sets.
Throughout all experiments, the testing data of size $M=2000$ are kept fixed to allow for fair comparisons.

For both networks, Tables~\ref{tab:denseNN-varyData-r2} and~\ref{tab:CNN-varyData-r2} show the lowest prediction errors for noise-free training and testing data, whereas the highest prediction errors occur when the training data do not contain noise but the test data do have noise.
For CNNs, predictions with noise in testing data improve significantly if the training data are also noisy (Table~\ref{tab:CNN-varyData-r2}, last column).  These results show the importance of training CNNs with noisy data in practice, even if noise-free simulated model outputs are available, because data from experiments or measurements are very likely polluted by some amount of noise.
Comparing the two network architectures, dense NN and CNN, the CNN delivers lower prediction errors in most cases except when the training of the CNN is performed with noise-free data while testing data have noise.  Only in this case, the dense NN demonstrates significantly better prediction skills (dense NN has $\Rsq>0.92$ vs.\ CNN has $\Rsq<0.90$) and therefore shows that dense layers are less sensitive to the presence or absence of noise in the training data.

We complete the discussion about prediction errors in the presence of observational noise by presenting scatterplots in Figures~\ref{fig:denseNN-varyNoise} and~\ref{fig:CNN-varyNoise} describing the distribution of errors for varying values of model parameters $\theta_0$ and $\theta_1$.
Each graph in the figures shows the true value of $\theta_0$ or $\theta_1$ from the test data set on the horizontal axis and the corresponding predicted value on the vertical axis.  Deviations of the predictions (blue dots) from the ideal (orange line) show the prediction errors.
The figures support the summarized statistics in Tables~\ref{tab:denseNN-varyData-r2} and~\ref{tab:CNN-varyData-r2} discussed above and additionally show that the spread of prediction errors is largely uniform across values of $\theta_0$ and $\theta_1$.  Slightly larger errors can appear at the lower and upper bounds of $\theta_0$ and $\theta_1$; these are more pronounced for the dense NN than for the CNN.

\paragraph{Simulations of FitzHugh--Nagumo model from predicted parameters}

\begin{figure}
  \begin{tikzpicture}
    \tikzset{
      errortext/.style={black,font={\footnotesize},align=left},
      graphlabel/.style={black,font={\footnotesize},align=center,rotate=90},
    }
    \node [anchor=west] (img) at (0,-0.3)
      {\includegraphics[width=0.75\columnwidth]{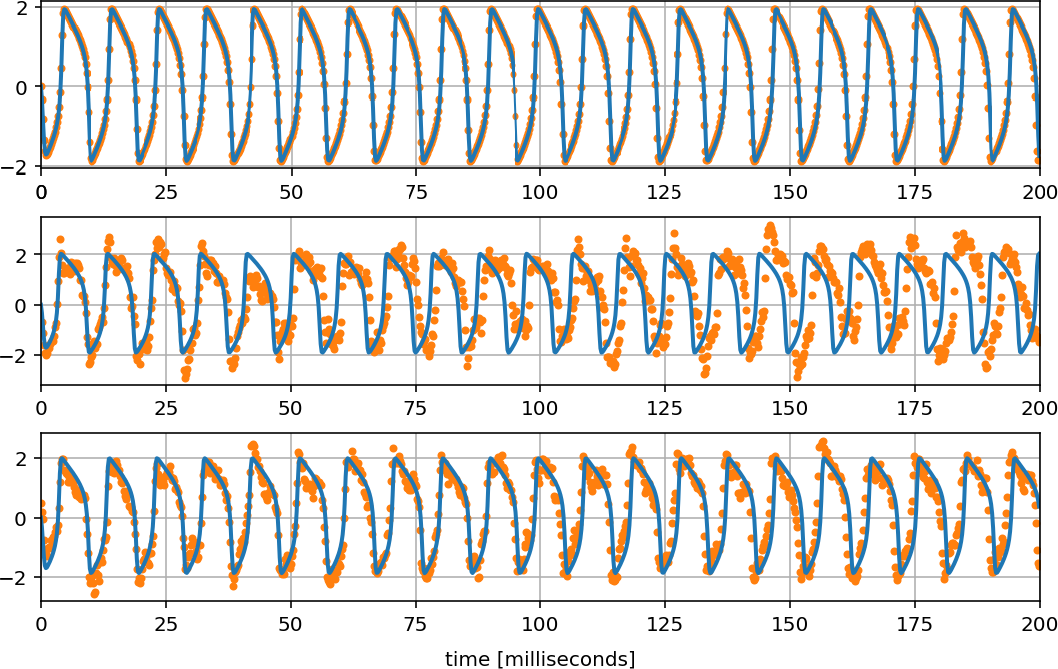}};
    \node [graphlabel,anchor=center] at (-0.3,+2.4)
      {train noise-free \\ test noise-free};
    \node [graphlabel,anchor=center] at (-0.3,0.0)
      {train noise-free \\ test with noise};
    \node [graphlabel,anchor=center] at (-0.3,-2.3)
      {train with noise \\ test with noise};
    \begin{scope}[xshift=0.75\columnwidth]
      \node [errortext,anchor=west] at (0.1,+2.4)
        {Time series error:\\
        $\text{Sq.\,bias} = \spp{2.8e-6}$\\
        $\text{C-MSE}     = \fpp{0.060849}$};
      \node [errortext,anchor=west] at (0.1,0.0)
        {Time series error:\\
        $\text{Sq.\,bias} = \spp{1.8e-3}$\\
        $\text{C-MSE}     = \fpppp{2.001331}$};
      \node [errortext,anchor=west] at (0.1,-2.3)
        {Time series error:\\
        $\text{Sq.\,bias} = \spp{1.2e-3}$\\
        $\text{C-MSE}     = \fppp{0.536042}$};
    \end{scope}
  \end{tikzpicture}
  \vspace{-5ex}
  \caption{Simulations of FitzHugh--Nagumo model (blue lines) using parameters %
  from CNN predictions; corresponding data that gave rise to prediction are shown as orange dots.
  Each graph shows the median percentile of MSE for the following cases.
  Top: training and testing data noise-free;
  Middle: training data noise-free and testing data with noise;
  Bottom: training and testing data with noise.}
  \label{fig:simODE-CNN-summary}
\end{figure}
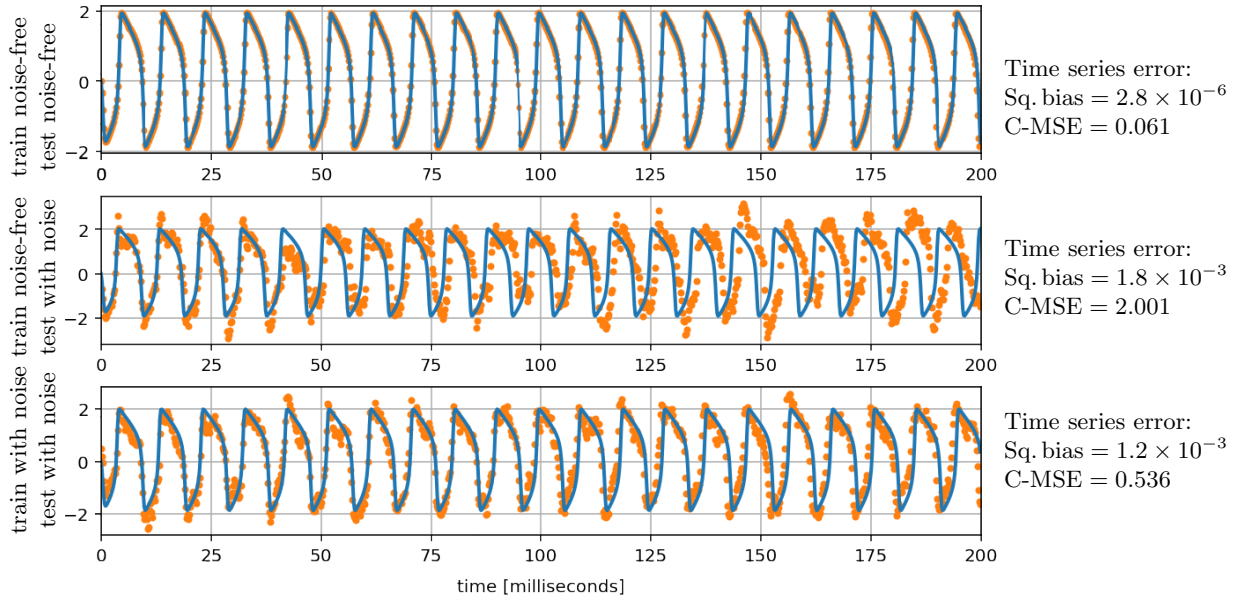

We extend the evaluation of results on model parameter predictions and now evaluate simulations of the FitzHugh--Nagumo model \eqref{eq:fhn} with predicted parameters.  The results presented here show the errors of the neural network predictions propagated through the forward problem.  We focus on the CNN with $n_f\times[1,2,4]$, $n_f=8$, and we consider the same three different cases of presence and/or absence of noise as before.
Figure~\ref{fig:simODE-CNN-summary} shows three FitzHugh--Nagumo simulations, in which the top graph represents the case where both training and testing data do not contain noise; in the middle graph only testing data contain noise; and the bottom graph shows results where both training and testing data contain noise.
Each of the three graphs is selected from the testing data set as the median of the MSE between true and predicted parameters.
More detailed results for FitzHugh--Nagumo simulations from predicted parameters are given in Appendix~\ref{sec:results-simulation}, where we show results for the 10\textsuperscript{th}, 25\textsuperscript{th}, 75\textsuperscript{th}, and 90\textsuperscript{th} percentiles of MSE between simulated and testing time series.

We observe a nearly optimal overlapping of simulated output from predicted parameters and test data corresponding to ``true'' parameters for the noise-free case (Figure~\ref{fig:simODE-CNN-summary}, top).
When noise is added to training or testing data, the simulated time series' show shifted spikes, which become more pronounced as simulation time increases (100--200 milliseconds).  The shifting of spikes over time represents discrepancies with respect to the frequency of a periodic time series.  This effect is more pronounced when the training data is noise free (Figure~\ref{fig:simODE-CNN-summary}, middle) and only mildly visible when the training data has noise (Figure~\ref{fig:simODE-CNN-summary}, bottom).
Overall, the results here and in Appendix~\ref{sec:results-simulation} quantify the accuracy and reliability of the proposed estimation in order to generate realistic solutions of the FitzHugh--Nagumo model from estimated parameters with convolutional networks.

\subsection{Joint estimation of FitzHugh--Nagumo parameters and noise parameters}
\label{sec:results-noise-prediction}

In addition to inferring parameters of the FitzHugh--Nagumo ODE, we are also interested in inferring properties of the noise contained in a time series.  Therefore, we target the simultaneous estimation of model and noise parameters with a single NN in this section.
This means that we extend the output space of the NN-based reconstruction map from two model-parameters $(\theta_0, \theta_1)$ to four parameters $(\theta_0, \theta_1, \sigma, \rho)$, where $\sigma$ is the noise standard deviation and $\rho$ is the autocorrelation parameter of the noise model~\eqref{eq:noise-model}.
Such a joint estimation of parameters of (deterministic) physical models based on differential equations and, simultaneously, of parameters of statistical models is challenging because of the extremely different time scales in the physical vs.\ noise stochastic processes.
Due to these difficulties, joint parameter estimation governed by physical and statistical models are rarely attempted in the literature.

\begin{table}
  \caption{%
    Median-APE (\Rsq) of joint parameter predictions of the ODE and noise parameters, using a CNN with $n_f\times[1,2,4]$, $n_f=8$.
    Providing Fourier spectrum to network becomes crucial when inferring noise parameters, yielding best results for time \& Fourier data.}
  \label{tab:CNN-noisePredict}
  \centering
  \vspace{1ex} %
  %
% CNN
% Med-APE (R^2)
%
\footnotesize
\begin{tabular}{clcccc}
  \toprule
  % table header
  \multirow{2}{*}{$N$} & \multirow{2}{*}{\thead{Data type}} & \multicolumn{2}{c}{\thead{FitzHugh--Nagumo parameter}} & \multicolumn{2}{c}{\thead{Noise parameter}} \\
                       &                                    & $\theta_0$ & $\theta_1$                                & $\sigma$ & $\rho$                           \\
  \midrule
  &Time            & \fppp{.1259816725406529} (\fppp{.8967960457188654}) & \fppp{.2300365690284891} (\fppp{.7784798446348100}) &\fppp{.1102501867679319} (\fpp{-.7626308414722736}) & \fpp{.0644409244005181} (\fpp{-.7865013801185115}) \\
  \thead{500}
  &Fourier         & \fppp{.2579907914826374} (\fppp{.4494779962905293}) & \fppp{.3485541073535537} (\fppp{.5186938864932531}) & \fpp{.0714760680632099} (\fppp{.4156178896175715}) & \fpp{.0324984477734047} (\fppp{.5389771822926841}) \\
  &Time \& Fourier & \fppp{.1030618306601233} (\fppp{.9211445957894405}) & \fppp{.2092448546388639} (\fppp{.7925879633066604}) & \fpp{.0627806439908616} (\fppp{.5491811335932968}) & \fpp{.0298915868857765} (\fppp{.6064865872371104}) \\
  \midrule
  &Time            & \fppp{.1154830064285569} (\fppp{.9140750658250013}) & \fppp{.2127025650484198} (\fppp{.8119365244772605}) &\fppp{.1128142029453091} (\fpp{-.6153832881292993}) & \fpp{.0625991713455814} (\fpp{-.7995185580044071}) \\
  \thead{1000}
  &Fourier         & \fppp{.2428024499594258} (\fppp{.4602846325591565}) & \fppp{.3149728960623374} (\fppp{.5769546507957903}) & \fpp{.0642052288936297} (\fppp{.5238639744231265}) & \fpp{.0282594050055054} (\fppp{.6025115189542501}) \\
  &Time \& Fourier & \fppp{.1027145855558811} (\fppp{.9345466433100864}) & \fppp{.1918758145229131} (\fppp{.8564322098707369}) & \fpp{.0576581718363855} (\fppp{.5891806981373909}) & \fpp{.0275079995034520} (\fppp{.6446162273980980}) \\
  \midrule
  &Time            & \fpp{.0891102031992249}  (\fppp{.9487972276204290}) & \fppp{.1684716772626455} (\fppp{.9066145289225189}) & \fpp{.0679178204021508} (\fppp{.4641529575200627}) & \fpp{.0500520322717141} (\fpp{-.1456121433752267}) \\
  \thead{4000}
  &Fourier         & \fppp{.2297922211572736} (\fppp{.5173295906763518}) & \fppp{.2587075849757031} (\fppp{.7479513424441617}) & \fpp{.0563190493443905} (\fppp{.6114556114929860}) & \fpp{.0267003567926474} (\fppp{.6533252426970374}) \\
  &Time \& Fourier & \fpp{.0764263731305400}  (\fppp{.9624861700668905}) & \fppp{.1363826136041256} (\fppp{.9156969386674416}) & \fpp{.0533900689238396} (\fppp{.6565859282612760}) & \fpp{.0257485643005741} (\fppp{.7001645387325215}) \\
  \midrule
  &Time            & \fpp{.0703546304003305}  (\fppp{.9615130500349991}) & \fppp{.1383522321024600} (\fppp{.9325790575745093}) & \fpp{.0576158214497447} (\fppp{.6265298715807888}) & \fpp{.0300291960140576} (\fppp{.5568505698797235}) \\
  \thead{8000}
  &Fourier         & \fppp{.1616920064146986} (\fppp{.5797300950368778}) & \fppp{.2146561764204512} (\fppp{.7966089274482634}) & \fpp{.0511845699387243} (\fppp{.6690450083143835}) & \fpp{.0234972567657431} (\fppp{.7213329393039570}) \\
  &Time \& Fourier & \fpp{.0659359002554415}  (\fppp{.9675096475244543}) & \fppp{.1097766933052231} (\fppp{.9419714248610168}) & \fpp{.0504459950356838} (\fppp{.6835126736501005}) & \fpp{.0237796441759850} (\fppp{.7217886018790838}) \\
  \bottomrule
\end{tabular}

\end{table}

We utilize the framework for computationally learning reconstruction maps  described in Section~\ref{sec:methods}.  Our framework requires samples for training and testing from a prior distribution; therefore we augment the prior for ODE model parameters $\theta_0, \theta_1$ from Equation~\eqref{eq:prior} with the prior information \eqref{eq:noise-prior} for the noise parameters $\sigma, \rho$.
In order to address the extreme range of time scales between ODE and noise models, we employ the Fourier transform of a noisy time series.  This yields three different types of observational data that can be used to train a network and perform predictions: a time series, a Fourier spectrum, and a time series combined with its Fourier spectrum. %
We consider the three data types as well as different sizes of training data sets in our numerical experiments for joint ODE and noise parameter estimation.

In practice, it is necessary to preprocess the different types of data with scaling transformations in order to rescale time series and Fourier transforms as well as the ODE and noise parameters; this can be carried out with standard libraries (e.g., scikit-learn).
The results are presented in Table~\ref{tab:CNN-noisePredict} for a CNN consisting of $3$ convolutional layers with $n_f\times[1,2,4]$, $n_f=8$ followed by $2$ dense layers.
The crucial role of the Fourier spectrum in order to estimate noise parameters is clearly demonstrated by negative \Rsq\ values where only time series data is provided to the CNN.  If spectrum data is omitted, very large amounts of training data are necessary ($N=8000$ in this case) in order to achieve a merely non-negative \Rsq\ for $\sigma,\rho$.
On the other hand, the Fourier spectrum by itself yields significantly larger errors for ODE parameters compared with time series data.
As a consequence, the best prediction performance for all four parameters is achieved when the CNN is trained with both a time series and the corresponding Fourier spectrum.
Overall, our framework for computationally learning NN-based reconstruction maps has demonstrated accurate predictions of model parameters, the reconstruction maps are able to handle noisy data very well, and previously challenging or even infeasible inference results become within reach as shown for the joint estimation of ODE and noise parameters.

\section{Conclusion}
\label{sec:conclusion}

In this paper, we build an estimation framework for an inverse problem governed by an ODE, the FitzHugh--Nagumo model.
The estimation consists of recovering model parameters as a prediction output from neural networks that are trained on time series solutions of the model as input.
Our study %
shows the efficacy of neural network-based parameter recovery
when traditional optimization techniques would fail to minimize the misfit function of the problem.
In particular, we propose and compare a range of different architectures of dense and convolutional NNs in order to computationally learn reconstruction maps of the inverse problem in different situations (ideal train and test data, noisy train and test data, and noisy test data only).
Prediction quality is carefully assessed through different statistical evaluation metrics and through the assessment of the FitzHugh--Nagumo model solutions calculated for the predicted model parameters.

CNN architectures mostly show the lowest errors when recovering parameters, which can be attributed to their locally acting kernels being advantageous for time-evolving data.
Additionally, CNNs extract crucial properties or dynamics of the ODE output when predicting parameters from arbitrarily chosen partial observations; dense NNs, in contrast, perform significantly worse.
In only one case can dense networks achieve better results than CNNs, which is when training is performed with noise-free data whereas prediction data are polluted with noise.  Thus, dense NNs demonstrate better generalization properties to testing data with ``unseen'' noise levels.
Our NN-based parameter estimation framework can generalize to other time dependent processes, because we successfully carry out a joint estimation of parameters from an ODE model and an autocorrelated noise model.

Future directions include embedding uncertainty quantification methods in NN-based parameter estimations. %
Uncertainty can be considered with respect to model and data discrepancies but also with respect to network architecture and trained NN parameters.
Some recent studies focus on loss functions and propose statistical losses that
incorporate additional information about the data, such as an correlation structure %
\citep{constantinescu2020}.
A different approach is to propose NN architectures for automatically selecting loss functions. \cite{aakesson2020convolutional} shows that loss functions for temporal statistical processes computed from a NN can effectively be used for inferring parameters within an ABC framework.
The present work makes use only of the mean squared error as a loss; therefore, one can explore the incorporation of prior information and a description of uncertainty in order to approximate posteriors of the inverse problem.
Our reconstruction maps rely on
training data generation
by solving the forward problem numerous times.  To mitigate
this computational cost,
one can explore reduced order model techniques, which recently have been advanced through machine learning methods \citep{QianKramerEtAl20, RoseberryVillaEtAl20}.

\section*{Acknowledgments}
The effort of Johann Rudi is based in part upon work supported by the U.S. Department of Energy, Office of Science, under contract DE-AC02-06CH11357.
The effort of Amanda Lenzi and Julie Bessac is based in part on work supported by the U.S. Department of Energy, Office of Science, Office of Advanced Scientific Computing Research (ASCR), under contract DE-AC02-06CH11347.

\bibliographystyle{plain}
\bibliography{references}

\begin{thebibliography}{10}

\bibitem{AdlerOktem17}
Jonas Adler and Ozan {\"O}ktem.
\newblock Solving ill-posed inverse problems using iterative deep neural
  networks.
\newblock {\em Inverse Problems}, 33(12):124007, 2017.

\bibitem{aakesson2020convolutional}
Mattias {\AA}kesson, Prashant Singh, Fredrik Wrede, and Andreas Hellander.
\newblock Convolutional neural networks as summary statistics for approximate
  {B}ayesian computation.
\newblock {\em arXiv preprint arXiv:2001.11760}, 2020.

\bibitem{AlonsoMarder19}
Leandro~M Alonso and Eve Marder.
\newblock Visualization of currents in neural models with similar behavior and
  different conductance densities.
\newblock {\em eLife}, 8:e42722, 2019.

\bibitem{ArnoldLloyd18}
Andrea Arnold and Alun~L Lloyd.
\newblock An approach to periodic, time-varying parameter estimation using
  nonlinear filtering.
\newblock {\em Inverse Problems}, 34(10):105005, 2018.

\bibitem{BallnusHugEtAl17}
Benjamin Ballnus, Sabine Hug, Kathrin Hatz, Linus G{\"o}rlitz, Jan Hasenauer,
  and Fabian~J Theis.
\newblock Comprehensive benchmarking of {M}arkov chain {M}onte {C}arlo methods
  for dynamical systems.
\newblock {\em BMC Systems Biology}, 11(64), 2017.

\bibitem{BuhrySaighiEtAl08}
Laure Buhry, Sylvain Saighi, Audrey Giremus, Eric Grivel, and Sylvie Renaud.
\newblock Parameter estimation of the {H}odgkin--{H}uxley model using
  metaheuristics: application to neuromimetic analog integrated circuits.
\newblock In {\em 2008 IEEE Biomedical Circuits and Systems Conference}, pages
  173--176. IEEE, 2008.

\bibitem{CalvettiSomersalo07}
Daniela Calvetti and Erkki Somersalo.
\newblock {\em Introduction to {B}ayesian Scientific Computing: Ten Lectures on
  Subjective Computing}, volume~2 of {\em Surveys and Tutorials in the Applied
  Mathematical Sciences}.
\newblock Springer-Verlag New York, 2007.

\bibitem{chon1997}
Ki~H Chon and Richard~J Cohen.
\newblock Linear and nonlinear {ARMA} model parameter estimation using an
  artificial neural network.
\newblock {\em IEEE transactions on biomedical engineering}, 44(3):168--174,
  1997.

\bibitem{constantinescu2020}
Emil~M Constantinescu, No{\'e}mi Petra, Julie Bessac, and Cosmin~G Petra.
\newblock Statistical treatment of inverse problems constrained by differential
  equations-based models with stochastic terms.
\newblock {\em SIAM/ASA Journal on Uncertainty Quantification}, 8(1):170--197,
  2020.

\bibitem{CranmerBrehmerLouppe20}
Kyle Cranmer, Johann Brehmer, and Gilles Louppe.
\newblock The frontier of simulation-based inference.
\newblock {\em Proceedings of the National Academy of Sciences},
  117(48):30055--30062, 2020.

\bibitem{creel2017neural}
Michael Creel.
\newblock Neural nets for indirect inference.
\newblock {\em Econometrics and Statistics}, 2:36--49, 2017.

\bibitem{CressieWikle15}
Noel Cressie and Christopher~K Wikle.
\newblock {\em Statistics for spatio-temporal data}.
\newblock John Wiley \& Sons, 2015.

\bibitem{DalyGavaghanEtAl18}
Aidan~C Daly, David Gavaghan, Jonathan Cooper, and Simon Tavener.
\newblock Inference-based assessment of parameter identifiability in nonlinear
  biological models.
\newblock {\em Journal of The Royal Society Interface}, 15(144):20180318, 2018.

\bibitem{DalyGavaghanEtAl15}
Aidan~C Daly, David~J Gavaghan, Chris Holmes, and Jonathan Cooper.
\newblock Hodgkin--{H}uxley revisited: reparametrization and identifiability
  analysis of the classic action potential model with approximate {B}ayesian
  methods.
\newblock {\em Royal Society open science}, 2(12):150499, 2015.

\bibitem{DengWangChe09}
Bin Deng, Jiang Wang, and Yenqiu Che.
\newblock A combined method to estimate parameters of neuron from a heavily
  noise-corrupted time series of active potential.
\newblock {\em Chaos: An Interdisciplinary Journal of Nonlinear Science},
  19(1):015105, 2009.

\bibitem{DoiOnodaKumagai02}
Shinji Doi, Yuichi Onoda, and Sadatoshi Kumagai.
\newblock Parameter estimation of various {H}odgkin--{H}uxley-type neuronal
  models using a gradient-descent learning method.
\newblock In {\em Proceedings of the 41st SICE Annual Conference. SICE 2002.},
  volume~3, pages 1685--1688. IEEE, 2002.

\bibitem{DorukAbosharb19}
Resat~Ozgur Doruk and Laila Abosharb.
\newblock Estimating the parameters of {F}itz{H}ugh--{N}agumo neurons from
  neural spiking data.
\newblock {\em Brain Sciences}, 9(12):364, 2019.

\bibitem{dua2011}
Vivek Dua.
\newblock An artificial neural network approximation based decomposition
  approach for parameter estimation of system of ordinary differential
  equations.
\newblock {\em Computers \& chemical engineering}, 35(3):545--553, 2011.

\bibitem{ElfwingUchibeDoya18}
Stefan Elfwing, Eiji Uchibe, and Kenji Doya.
\newblock Sigmoid-weighted linear units for neural network function
  approximation in reinforcement learning.
\newblock {\em Neural Networks}, 107:3--11, 2018.

\bibitem{FanBohorquezYing19}
Yuwei Fan, Cindy~Orozco Bohorquez, and Lexing Ying.
\newblock {BCR}-{N}et: A neural network based on the nonstandard wavelet form.
\newblock {\em Journal of Computational Physics}, 384:1--15, 2019.

\bibitem{Fitzhugh61}
Richard FitzHugh.
\newblock Impulses and physiological states in theoretical models of nerve
  membrane.
\newblock {\em Biophysical Journal}, 1(6):445--466, 1961.

\bibitem{GoncalvesLueckmannEtAl20}
Pedro~J Gon{\c{c}}alves, Jan-Matthis Lueckmann, Michael Deistler, Marcel
  Nonnenmacher, Kaan {\"O}cal, Giacomo Bassetto, Chaitanya Chintaluri,
  William~F Podlaski, Sara~A Haddad, Tim~P Vogels, David~S Greenberg, and
  Jakob~H Macke.
\newblock Training deep neural density estimators to identify mechanistic
  models of neural dynamics.
\newblock {\em eLife}, 9:e56261, 2020.

\bibitem{Goodfellow2016}
Ian Goodfellow, Yoshua Bengio, and Aaron Courville.
\newblock {\em Deep Learning}.
\newblock MIT Press, 2016.

\bibitem{GutenkunstWaterfallEtAl07}
Ryan~N Gutenkunst, Joshua~J Waterfall, Fergal~P Casey, Kevin~S Brown,
  Christopher~R Myers, and James~P Sethna.
\newblock Universally sloppy parameter sensitivities in systems biology models.
\newblock {\em PLOS Computational Biology}, 3(10):e189, 2007.

\bibitem{HamiltonBerrySauer18}
Franz Hamilton, Tyrus Berry, and Timothy Sauer.
\newblock Tracking intracellular dynamics through extracellular measurements.
\newblock {\em PloS one}, 13(10):e0205031, 2018.

\bibitem{HayouDoucetRousseau2018}
Soufiane Hayou, Arnaud Doucet, and Judith Rousseau.
\newblock On the selection of initialization and activation function for deep
  neural networks.
\newblock {\em arXiv preprint arXiv:1805.08266}, 2018.

\bibitem{HodgkinHuxley52}
Alan~L Hodgkin and Andrew~F Huxley.
\newblock A quantitative description of membrane current and its application to
  conduction and excitation in nerve.
\newblock {\em The Journal of physiology}, 117(4):500--544, 1952.

\bibitem{jiang2017learning}
Bai Jiang, Tung-yu Wu, Charles Zheng, and Wing~H. Wong.
\newblock Learning summary statistic for approximate {B}ayesian computation via
  deep neural network.
\newblock {\em Statistica Sinica}, pages 1595--1618, 2017.

\bibitem{JoliveRauchEtAlt06}
Renaud Jolivet, Alexander Rauch, Hans-Rudolf L{\"u}scher, and Wulfram Gerstner.
\newblock Predicting spike timing of neocortical pyramidal neurons by simple
  threshold models.
\newblock {\em Journal of Computational Neuroscience}, 21(1):35--49, 2006.

\bibitem{KaipioSomersalo05}
Jari Kaipio and Erkki Somersalo.
\newblock {\em Statistical and Computational Inverse Problems}, volume 160 of
  {\em Applied Mathematical Sciences}.
\newblock Springer-Verlag New York, 2005.

\bibitem{KhooYing19}
Yuehaw Khoo and Lexing Ying.
\newblock Switch{N}et: A neural network model for forward and inverse
  scattering problems.
\newblock {\em SIAM Journal on Scientific Computing}, 41(5):A3182--A3201, 2019.

\bibitem{KingmaBa15}
Diederik~P Kingma and Jimmy Ba.
\newblock Adam: A method for stochastic optimization.
\newblock In {\em International Conference for Learning Representation}, 2015.

\bibitem{kohavi1995study}
Ron Kohavi et~al.
\newblock A study of cross-validation and bootstrap for accuracy estimation and
  model selection.
\newblock In {\em Ijcai}, volume~14, pages 1137--1145. Montreal, Canada, 1995.

\bibitem{KrizhevskySutskeverHinton12}
Alex Krizhevsky, Ilya Sutskever, and Geoffrey~E Hinton.
\newblock {ImageNet} classification with deep convolutional neural networks.
\newblock In {\em Advances in Neural Information Processing Systems},
  volume~25, pages 1097--1105. Curran Associates, Inc., 2012.

\bibitem{lecun1999}
Yann LeCun, Patrick Haffner, L{\'e}on Bottou, and Yoshua Bengio.
\newblock Object recognition with gradient-based learning.
\newblock In {\em Shape, contour and grouping in computer vision}, pages
  319--345. Springer, 1999.

\bibitem{mills1991time}
Terence~C Mills.
\newblock {\em Time Series Techniques for Economists}.
\newblock Cambridge University Press, 1991.

\bibitem{morshed1998}
Jahangir Morshed and Jagath~J. Kaluarachchi.
\newblock Parameter estimation using artificial neural network and genetic
  algorithm for free-product migration and recovery.
\newblock {\em Water Resources Research}, 34(5):1101--1113, 1998.

\bibitem{NagumoArimotoYoshizawa62}
Jinichi Nagumo, Suguru Arimoto, and Shuji Yoshizawa.
\newblock An active pulse transmission line simulating nerve axon.
\newblock {\em Proceedings of the IRE}, 50(10):2061--2070, 1962.

\bibitem{NaudBathellierGerstner14}
Richard Naud, Brice Bathellier, and Wulfram Gerstner.
\newblock Spike-timing prediction in cortical neurons with active dendrites.
\newblock {\em Frontiers in Computational Neuroscience}, 8:90, 2014.

\bibitem{RoseberryVillaEtAl20}
Thomas O'Leary-Roseberry, Umberto Villa, Peng Chen, and Omar Ghattas.
\newblock Derivative-informed projected neural networks for high-dimensional
  parametric maps governed by {PDE}s.
\newblock {\em arXiv preprint arXiv:2011.15110}, 2020.

\bibitem{PaganiManzoniQuarteroni17}
Stefano Pagani, Andrea Manzoni, and Alfio Quarteroni.
\newblock Efficient state/parameter estimation in nonlinear unsteady {PDE}s by
  a reduced basis ensemble {K}alman filter.
\newblock {\em SIAM/ASA Journal on Uncertainty Quantification}, 5(1):890--921,
  2017.

\bibitem{parikh2020}
Jaimit Parikh, James Kozloski, and Viatcheslav Gurev.
\newblock Integration of {AI} and mechanistic modeling in generative
  adversarial networks for stochastic inverse problems.
\newblock {\em arXiv preprint arXiv:2009.08267}, 2020.

\bibitem{PrinzBucherMarder04}
Astrid~A Prinz, Dirk Bucher, and Eve Marder.
\newblock Similar network activity from disparate circuit parameters.
\newblock {\em Nature Neuroscience}, 7(12):1345--1352, 2004.

\bibitem{QianKramerEtAl20}
Elizabeth Qian, Boris Kramer, Benjamin Peherstorfer, and Karen Willcox.
\newblock Lift \& learn: Physics-informed machine learning for large-scale
  nonlinear dynamical systems.
\newblock {\em Physica D: Nonlinear Phenomena}, 406:132401, 2020.

\bibitem{radev2020}
Stefan~T. Radev, Ulf~K. Mertens, Andreass Voss, Lynton Ardizzone, and Ullrich
  K{\"o}the.
\newblock {BayesFlow}: Learning complex stochastic models with invertible
  neural networks.
\newblock {\em arXiv preprint arXiv:2003.06281}, 2020.

\bibitem{schmidhuber2015}
J{\"u}rgen Schmidhuber.
\newblock Deep learning in neural networks: An overview.
\newblock {\em Neural networks}, 61:85--117, 2015.

\bibitem{Stuart10}
Andrew~M. Stuart.
\newblock Inverse problems: {A B}ayesian perspective.
\newblock {\em Acta Numerica}, 19:451--559, 2010.

\bibitem{Tarantola05}
Albert Tarantola.
\newblock {\em Inverse Problem Theory and Methods for Model Parameter
  Estimation}.
\newblock SIAM, Philadelphia, PA, 2005.

\bibitem{taylor2001}
Karl~E Taylor.
\newblock Summarizing multiple aspects of model performance in a single
  diagram.
\newblock {\em Journal of Geophysical Research: Atmospheres},
  106(D7):7183--7192, 2001.

\bibitem{VanGeitEtAl08}
Werner Van~Geit, Erik De~Schutter, and Pablo Achard.
\newblock Automated neuron model optimization techniques: a review.
\newblock {\em Biological Cybernetics}, 99:241--251, 2008.

\end{thebibliography}

\appendix

\section{Influence of parameters on FitzHugh–Nagumo model solutions and distribution of parameter samples}
\label{sec:parameters}

This section provides background information on how the inference parameters $\boldsymbol{\theta} \coloneqq (\theta_0, \theta_1)$ affect solutions of the FitzHugh--Nagumo ODE and thus the observational data that is used to generate the NN-based reconstruction maps.
Furthermore, the section discusses the distribution of parameter samples that are used to train the networks and test their predictions.

\begin{figure}
  \centering
  \includegraphics[width=0.48\columnwidth]{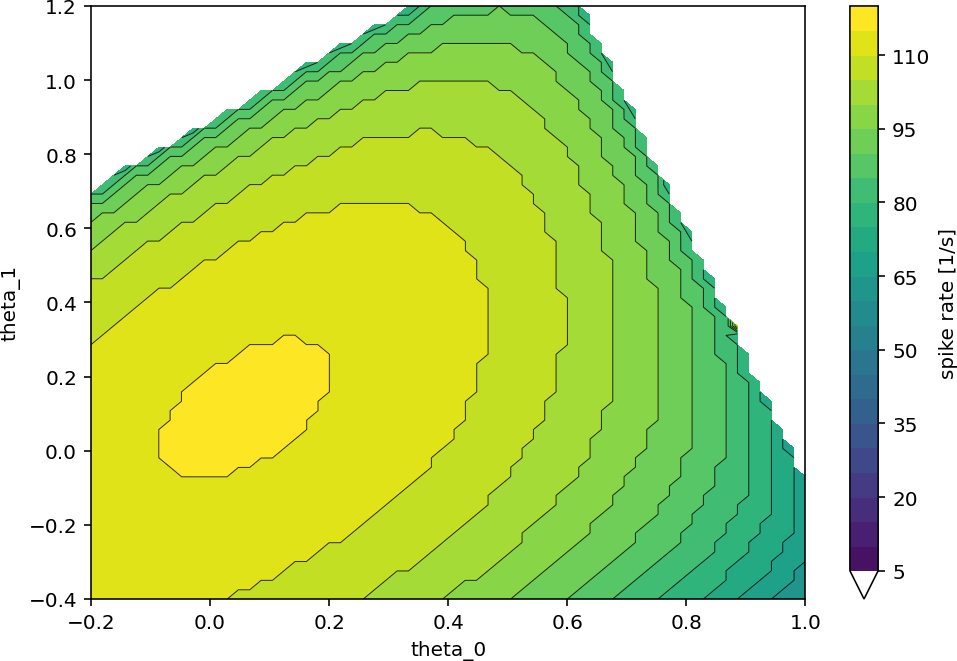}
  \hfill
  \includegraphics[width=0.48\columnwidth]{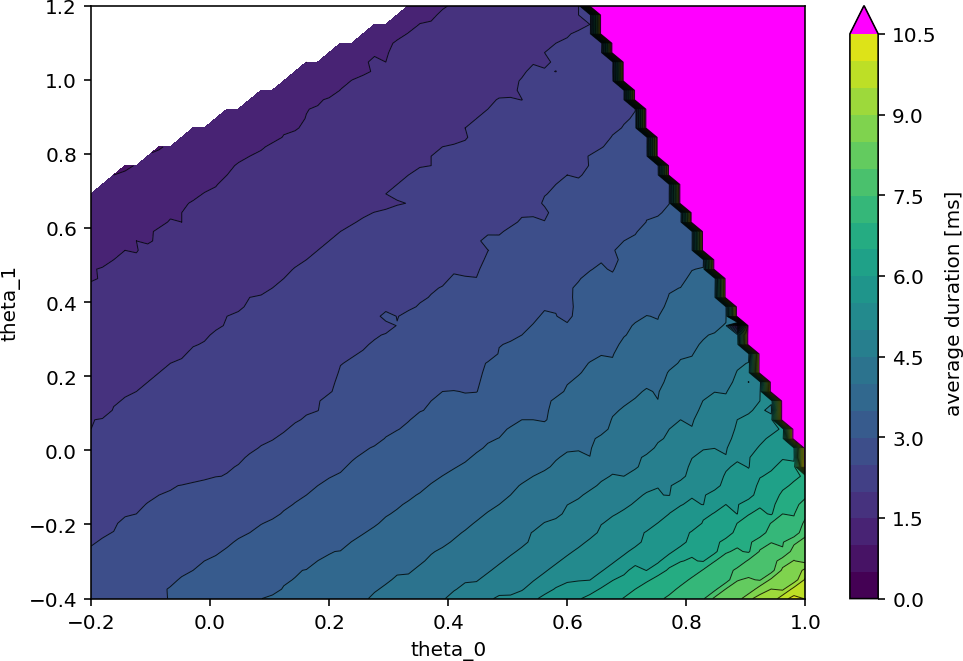}
  \vspace{-2ex}
  \caption{%
    Left: Spike rate of the membrane potential $u(t)$ stemming from solutions of the FitzHugh--Nagumo ODE for varying parameters $\theta_0$ and $\theta_1$, which are on the horizontal and vertical axes, respectively.
    Right: Average duration of spikes of the membrane potential.
    The triangular region on the top right in both plots (white color in left, magenta color in right plot) has only a single spike and $u(t)$ stays flat above the spike threshold of $1.5$.
    The triangular region on the top left (white color in both plots), on the other hand, has zero spikes.
  }
  \label{fig:spike-rate-duration}
  \vspace{4ex}
  \centering
  \includegraphics[width=0.48\columnwidth]{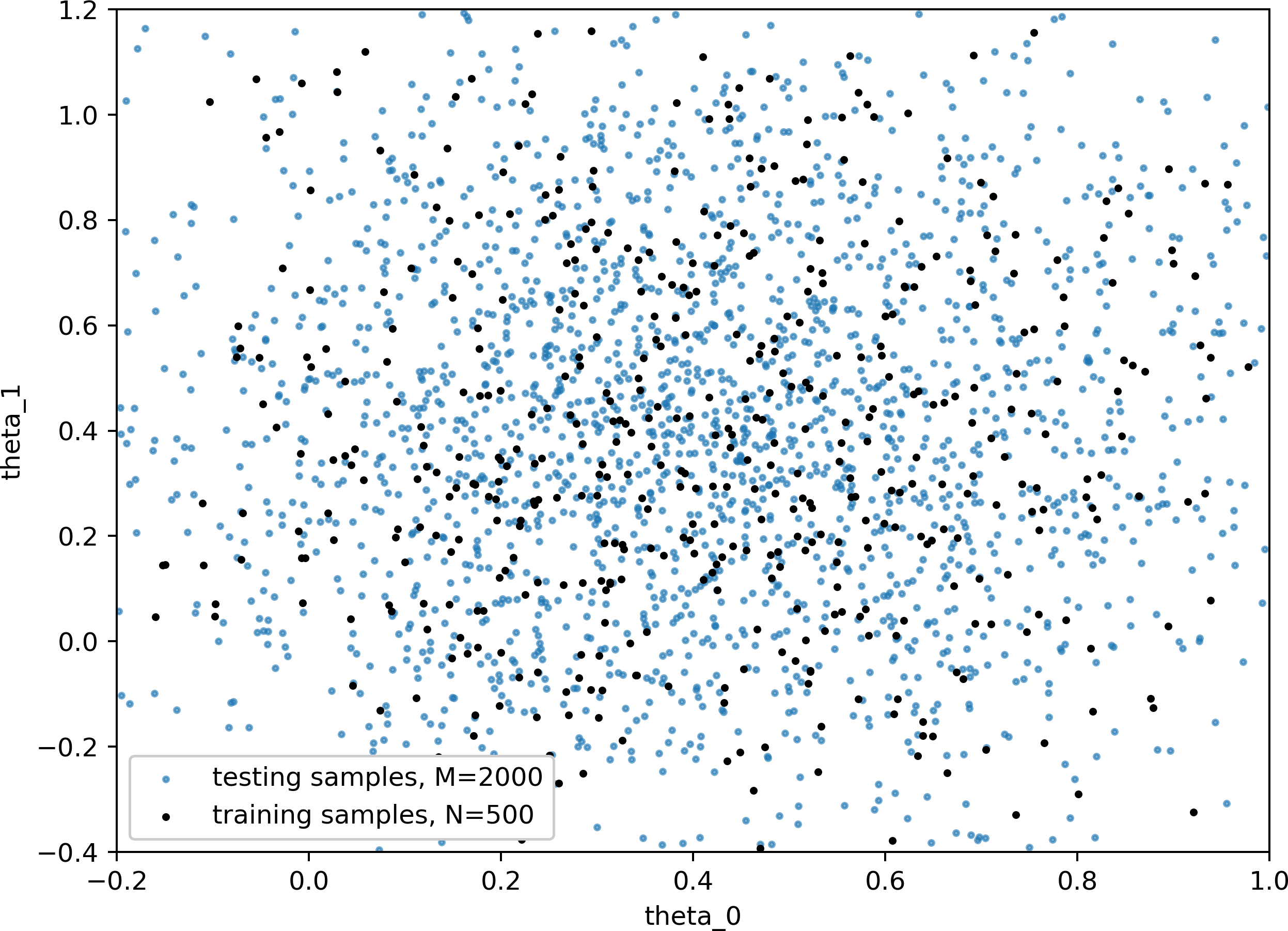}
  \hfill
  \includegraphics[width=0.48\columnwidth]{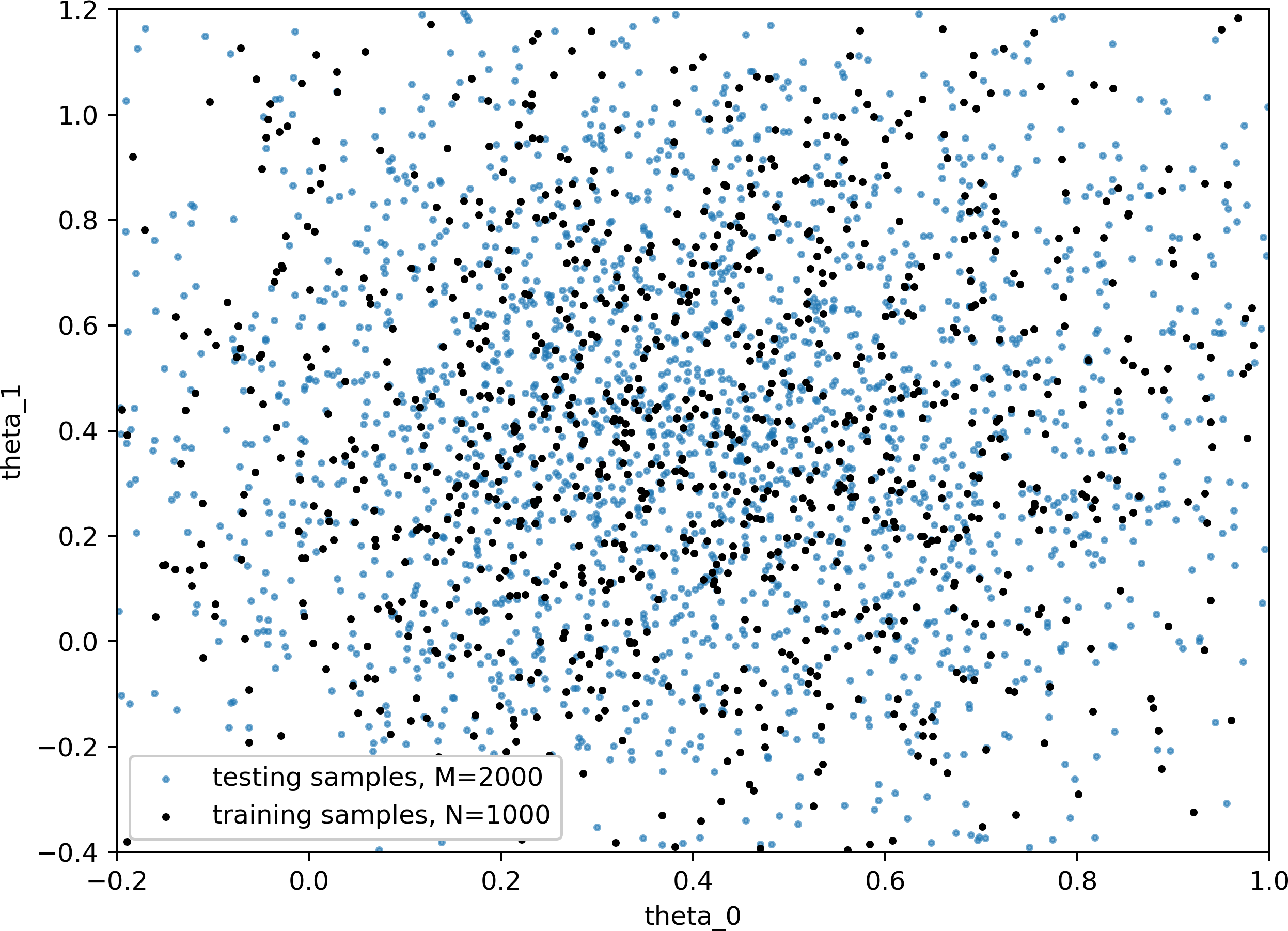}
  \vskip 1ex
  \includegraphics[width=0.48\columnwidth]{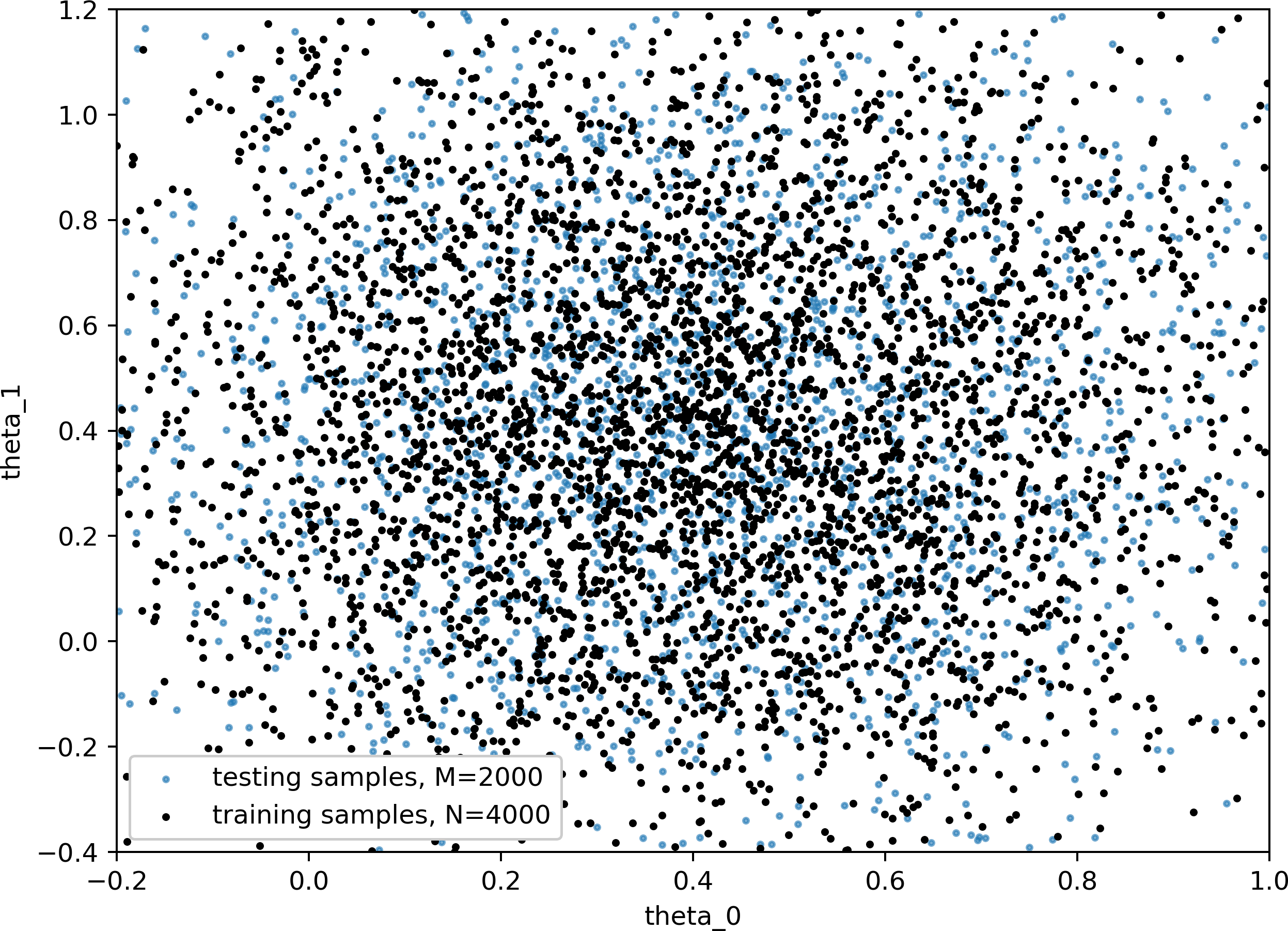}
  \hfill
  \includegraphics[width=0.48\columnwidth]{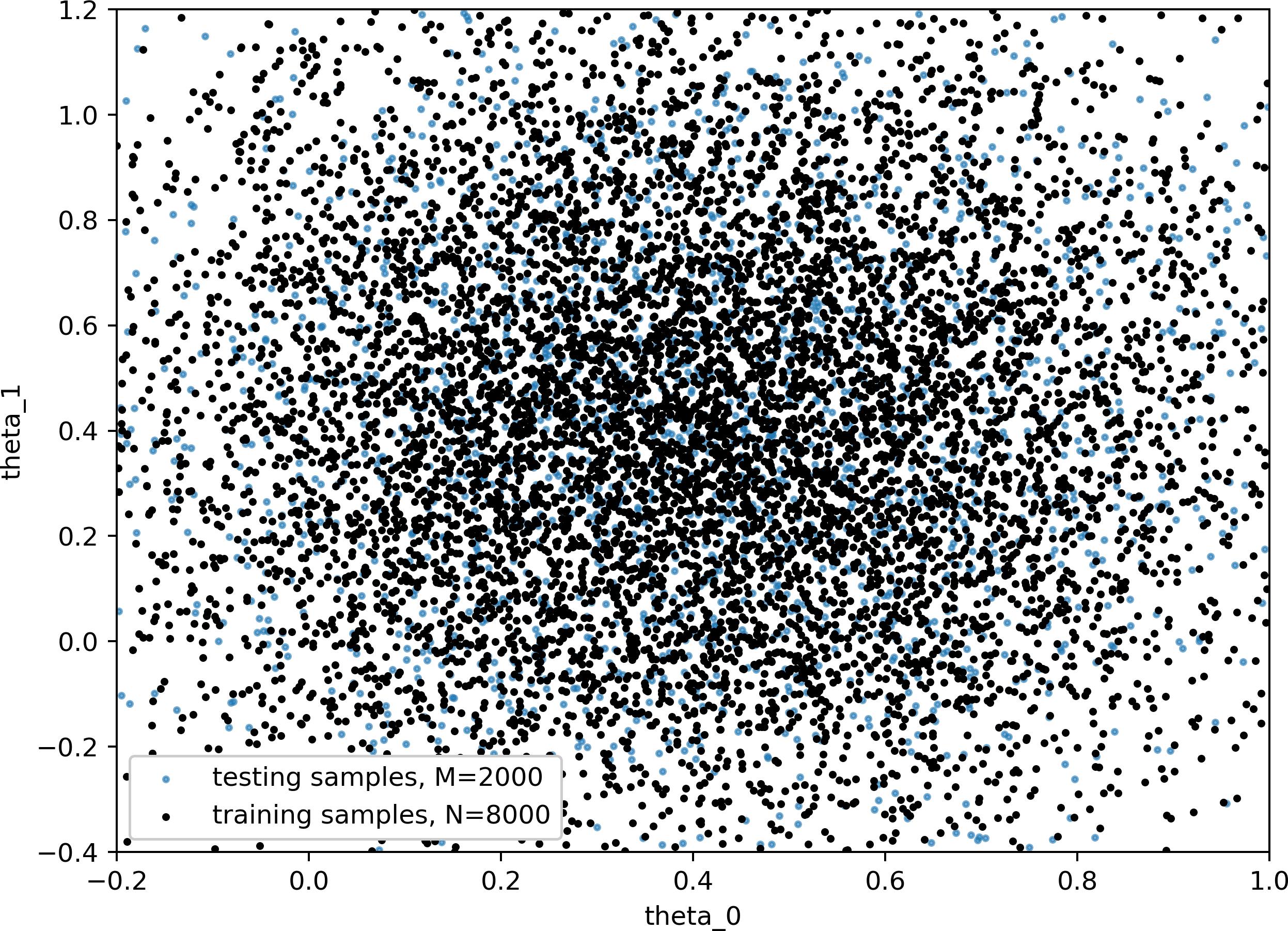}
  \vspace{-2ex}
  \caption{%
    Model parameters of testing set (blue dots) stay fixed ($M=2000$) while the size of the training set (black dots) increases ($N=500,1000,4000,8000$), clockwise from the top left scatter plot.
  }
  \label{fig:parameters}
\end{figure}

\paragraph{Influence of parameters on FitzHugh–Nagumo model solutions}
The choice of parameters $\boldsymbol{\theta} \coloneqq (\theta_0, \theta_1)$ for inference from the FitzHugh--Nagumo model is motivated by how $\boldsymbol{\theta}$ influences solutions of the ODE \eqref{eq:fhn}, specifically the membrane potential $u(t)$.
Two important characteristics of the oscillating membrane potential are the spike rate and spike duration.  Therefore, we visualize these quantities in Figure~\ref{fig:spike-rate-duration}.
In order to detect a spike of $u(t)$, we consider a value called spike threshold $u_\text{spike}$, which is a constant that determines the occurrence of a spike at time $t_\text{spike}$, if
  $u_\text{spike} \le u(t_\text{spike})$
and $u_\text{spike}>u(t_\text{spike}-\epsilon)$ for some $\epsilon>0$.
The spike duration is defined as the time from $t_\text{spike}$ until $u_\text{spike}>u(t)$ for $t>t_\text{spike}$.
To generate the plots in Figure~\ref{fig:spike-rate-duration}, we calculate the spike rate (i.e., number of spikes divided by time series length), which is depicted left in Figure~\ref{fig:spike-rate-duration}, and take the average duration over all spike durations to obtain the right plot in Figure~\ref{fig:spike-rate-duration}.
We observe in this figure that both the spike rate and the duration are controlled by $\theta_0$ and $\theta_1$ in nonlinear ways, and that the dependencies of spike rate and duration on the parameters are distinct from one another.
Additionally, we observe that some combinations of parameters generate zero or only one spike (flat purple, yellow, and white triangular regions in Figure~\ref{fig:spike-rate-duration}).  The NN-based reconstruction maps, however, have shown to predict parameters even from these more challenging time series.

\paragraph{Distribution of parameter samples for training and testing of neural networks}
Training and testing data for NNs are obtained from, first, sampling parameters $\boldsymbol{\theta}$ from their prior distributions \eqref{eq:prior} and discarding samples outside of prior bounds \eqref{eq:bounds} and, second, solving ODE \eqref{eq:fhn} and storing the membrane potential $u_{\boldsymbol{\theta}}$.
Figure~\ref{fig:parameters} presents the distribution of prior samples to illustrate how close training and testing samples of $\boldsymbol{\theta}$ are to one another.
The testing samples of size $M=2000$ are displayed as blue dots and are fixed in all four plots.  Fixing the testing samples ensures that the evaluation of prediction errors is comparable across all numerical experiments.
The training samples are varied in size $N=500, 1000, 4000, 8000$ and are shown as black dots in Figure~\ref{fig:parameters} (clockwise from top left scatter plot).  The configurations with $N=500, 1000$ training samples are further apart from each other, hence permitting blue dotted testing samples to be distinct from training sets.  The configurations with $N=4000, 8000$, on the other hand, exhibit dense training sets and cover most of the parameter space.

\section{K-fold cross-validation of selected neural network architectures and their sensitivity to the initialization of weights}
\label{sec:crossvalidation}

We assess further the quality of the selected NN along with the effects of different initial initializations of the weights in fitted dense NNs and CNNs using a $k$-fold cross-validation, with $k=6$.
After splitting the data set into six roughly equal-sized parts, the NNs are fit to five splits concatenated into a single training set and evaluation metrics are calculated when predicting the remaining sixth part of the data. We repeat this procedure to each of the six data splits, and a final mean and standard deviation of the prediction error are obtained by combining the estimates from the six splits that have been left out from training sets. A 5- to 10-fold cross-validation is typically recommended \citep{kohavi1995study}, and we find that six folds provide a good compromise between bias and variance for our case study.

Tables~\ref{tab:crossvalidation-denseNN-N1000} and~\ref{tab:crossvalidation-CNN-N1000} show the mean and standard deviation from the 6-fold cross-validation of the prediction skill scores outlined in Section~\ref{sec:metrics}. Each row of a table corresponds to a different seed when initializing the weights of the dense NNs and CNNs. We notice that the random seeds have little effect on the predictions, indicating that the method is robust to the randomness incurred in the fit of the networks.
Moreover, the lower scores from the CNN (Table~\ref{tab:crossvalidation-CNN-N1000}) compared to the dense NN (Table~\ref{tab:crossvalidation-denseNN-N1000}) substances the conclusions in Section~\ref{sec:results-exploration} that the CNN results in better predictions than the dense NN.

\begin{table}
  \caption{%
    6-fold cross-validation with (noise-free) training and testing sets of sizes $N=1000$ and $M=200$, respectively;
    random seeds for initializing network weights vary per row;
    using a \textbf{dense NN} with $4$ layers, $n_u=32$.}
  \label{tab:crossvalidation-denseNN-N1000}
  \centering
  \vspace{1ex} %
  %
% dense NN
%
\footnotesize
\setlength{\tabcolsep}{0.4em}  % set horizontal cell padding
\begin{tabular}{lccccccccccc}
  \toprule
  % table header
  \thead{Random seed} &
  \multicolumn{2}{c}{\thead{Squared bias}} &&
  \multicolumn{2}{c}{\thead{C-MSE}}        &&
  \multicolumn{2}{c}{\thead{Median-APE}}   &&
  \multicolumn{2}{c}{\thead{\Rsq}}
  \\
  &
  mean & std. &&
  mean & std. &&
  mean & std. &&
  mean & std.
  \\
  \midrule
  % table body
  seed \#1 & \fppp{6.04163522e-05} & \fppp{5.44986615e-05} && \fppp{2.95131304e-03} & \fppp{1.13449298e-03} && \fppp{0.0350468538} & \fpp{0.00806162094} && \fppp{0.970426910} & \fppp{0.0105257669} \\
  seed \#2 & \fppp{5.60503882e-05} & \fppp{6.27946344e-05} && \fppp{2.45288263e-03} & \fppp{9.87096037e-04} && \fppp{0.0269388745} & \fppp{0.0104133348} && \fppp{0.975850521} & \fpp{0.00797547721} \\
  seed \#3 & \fppp{3.78035546e-05} & \fppp{2.40419657e-05} && \fppp{2.40402848e-03} & \fppp{8.88077511e-04} && \fppp{0.0262819328} & \fpp{0.00633251461} && \fppp{0.975869831} & \fpp{0.00748275783} \\
  seed \#4 & \fppp{8.10987077e-05} & \fppp{5.81403754e-05} && \fppp{2.71873263e-03} & \fppp{1.25288903e-03} && \fppp{0.0281315401} & \fpp{0.00464780617} && \fppp{0.973239231} & \fppp{0.0101341892} \\
  seed \#5 & \fppp{5.33511522e-05} & \fppp{4.31804914e-05} && \fppp{2.31000146e-03} & \fppp{4.97546740e-04} && \fppp{0.0281562060} & \fpp{0.00547264980} && \fppp{0.976690378} & \fpp{0.00422129198} \\
  seed \#6 & \fppp{9.75675927e-05} & \fppp{1.48585735e-04} && \fppp{2.34695168e-03} & \fppp{6.26710872e-04} && \fppp{0.0314463969} & \fpp{0.00470602424} && \fppp{0.975575135} & \fpp{0.00530858138} \\
  seed \#7 & \fppp{5.54911017e-05} & \fppp{7.30190657e-05} && \fppp{2.34853961e-03} & \fppp{8.61789086e-04} && \fppp{0.0270845747} & \fpp{0.00686973894} && \fppp{0.976694882} & \fpp{0.00657901608} \\
  seed \#8 & \fppp{9.43730272e-05} & \fppp{7.90584836e-05} && \fppp{2.51767237e-03} & \fppp{6.28032711e-04} && \fppp{0.0328725951} & \fppp{0.0101765890} && \fppp{0.974272014} & \fpp{0.00466162871} \\
  seed \#9 & \fppp{3.81233340e-05} & \fppp{4.21377219e-05} && \fppp{2.16864509e-03} & \fppp{7.98302907e-04} && \fppp{0.0268950421} & \fpp{0.00525868925} && \fppp{0.977872317} & \fpp{0.00633146291} \\
  seed \#10& \fppp{5.08989476e-05} & \fppp{4.53351632e-05} && \fppp{2.48832141e-03} & \fppp{6.40554802e-04} && \fppp{0.0294561694} & \fpp{0.00587844817} && \fppp{0.975075789} & \fpp{0.00370541265} \\
  \bottomrule
\end{tabular}
  \vspace{4ex}
  \caption{%
    6-fold cross-validation with (noise-free) training and testing sets of sizes $N=1000$ and $M=200$, respectively;
    random seeds for initializing network weights vary per row;
    using a \textbf{CNN} with $n_f\times[1,2,4]$, $n_f=8$.}
  \label{tab:crossvalidation-CNN-N1000}
  \centering
  \vspace{1ex} %
  %
% CNN
%
\footnotesize
\setlength{\tabcolsep}{0.4em}  % set horizontal cell padding
\begin{tabular}{lccccccccccc}
  \toprule
  % table header
  \thead{Random seed} &
  \multicolumn{2}{c}{\thead{Squared bias}} &&
  \multicolumn{2}{c}{\thead{C-MSE}}        &&
  \multicolumn{2}{c}{\thead{Median-APE}}   &&
  \multicolumn{2}{c}{\thead{\Rsq}}
  \\
  &
  mean & std. &&
  mean & std. &&
  mean & std. &&
  mean & std.
  \\
  \midrule
  % table body
  seed \#1 & \fppp{3.04719219e-05} & \fppp{3.04603777e-05} && \fppp{5.39874764e-04} & \fppp{3.01544561e-04} && \fppp{0.0180835620} & \fpp{0.00376614997} && \fppp{0.993883583} & \fpp{0.00315838583} \\
  seed \#2 & \fppp{3.59448270e-05} & \fppp{3.99675100e-05} && \fppp{4.88196131e-04} & \fppp{2.60170872e-04} && \fppp{0.0165793699} & \fpp{0.00777368998} && \fppp{0.994260424} & \fpp{0.00337763803} \\
  seed \#3 & \fppp{2.87438703e-05} & \fppp{3.58750940e-05} && \fppp{4.48298779e-04} & \fppp{2.49516520e-04} && \fppp{0.0131128291} & \fpp{0.00653577764} && \fppp{0.994801691} & \fpp{0.00292725068} \\
  seed \#4 & \fppp{5.43101209e-05} & \fppp{4.10224891e-05} && \fppp{4.60349648e-04} & \fppp{2.10802304e-04} && \fppp{0.0175641730} & \fpp{0.00535287229} && \fppp{0.994329983} & \fpp{0.00253312648} \\
  seed \#5 & \fppp{3.60158710e-05} & \fppp{3.57671844e-05} && \fppp{5.39857425e-04} & \fppp{3.21079443e-04} && \fppp{0.0191087667} & \fpp{0.00356375949} && \fppp{0.993573096} & \fpp{0.00363571771} \\
  seed \#6 & \fppp{2.24093514e-05} & \fppp{2.73726130e-05} && \fppp{5.00885814e-04} & \fppp{2.77054355e-04} && \fppp{0.0169347351} & \fpp{0.00657534324} && \fppp{0.994198780} & \fpp{0.00338244852} \\
  seed \#7 & \fppp{4.77129942e-05} & \fppp{3.85437821e-05} && \fppp{4.91987530e-04} & \fppp{2.77192860e-04} && \fppp{0.0166632365} & \fpp{0.00472128921} && \fppp{0.994094706} & \fpp{0.00339490709} \\
  seed \#8 & \fppp{2.83165106e-05} & \fppp{2.70846236e-05} && \fppp{5.11637074e-04} & \fppp{1.77483767e-04} && \fppp{0.0139590238} & \fpp{0.00582949288} && \fppp{0.993938208} & \fpp{0.00223090956} \\
  seed \#9 & \fppp{1.12956840e-05} & \fppp{9.91729215e-06} && \fppp{4.47280524e-04} & \fppp{2.67145842e-04} && \fppp{0.0960608051} & \fpp{0.00351135705} && \fppp{0.994875988} & \fpp{0.00323630302} \\
  seed \#10& \fppp{2.33763314e-05} & \fppp{1.34813655e-05} && \fppp{4.83559639e-04} & \fppp{2.62047839e-04} && \fppp{0.0186085286} & \fpp{0.00617480556} && \fppp{0.994464158} & \fpp{0.00312187454} \\
  \bottomrule
\end{tabular}
\end{table}

\section{Extended results for simulation of FitzHugh--Nagumo model from predicted parameters}
\label{sec:results-simulation}

We support the observations from the last paragraph of Section~\ref{sec:results-noise} here by showing a wider scope of prediction errors.  Recall that Section~\ref{sec:results-noise} considers the evaluation of parameter prediction errors from neural networks by means of comparing simulations of the FitzHugh--Nagumo model from predicted parameters with the time series used as input to the NN.  The predictions are carried out by the CNN selected in Section~\ref{sec:results-exploration} with $n_f\times[1,2,4]$, $n_f=8$.

\begin{figure}
  \centering
  \includegraphics[width=0.96\columnwidth]{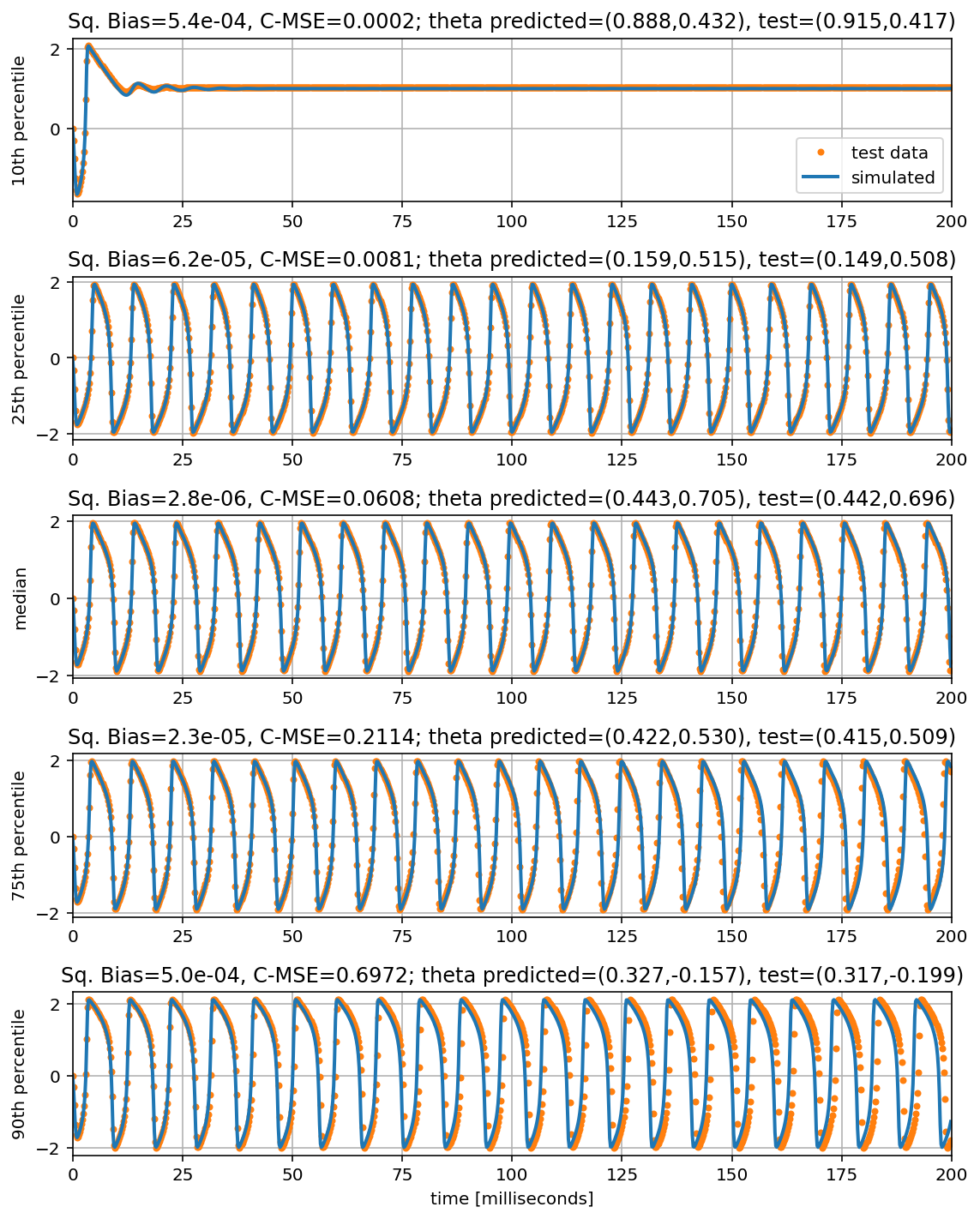}
  \caption{Simulations of FitzHugh--Nagumo model (blue lines) using parameters %
    from CNN predictions; corresponding data that gave rise to prediction are shown as orange dots.
    Training samples ($N=1000$) and testing samples ($M=2000$) are both noise-free.
    Graphs from top to bottom show increasing quantiles of MSE between true and estimated time series.}
  \label{fig:simODE-CNN-noiseTrain0-noiseTest0}
\end{figure}

\begin{figure}
  \centering
  \includegraphics[width=0.96\columnwidth]{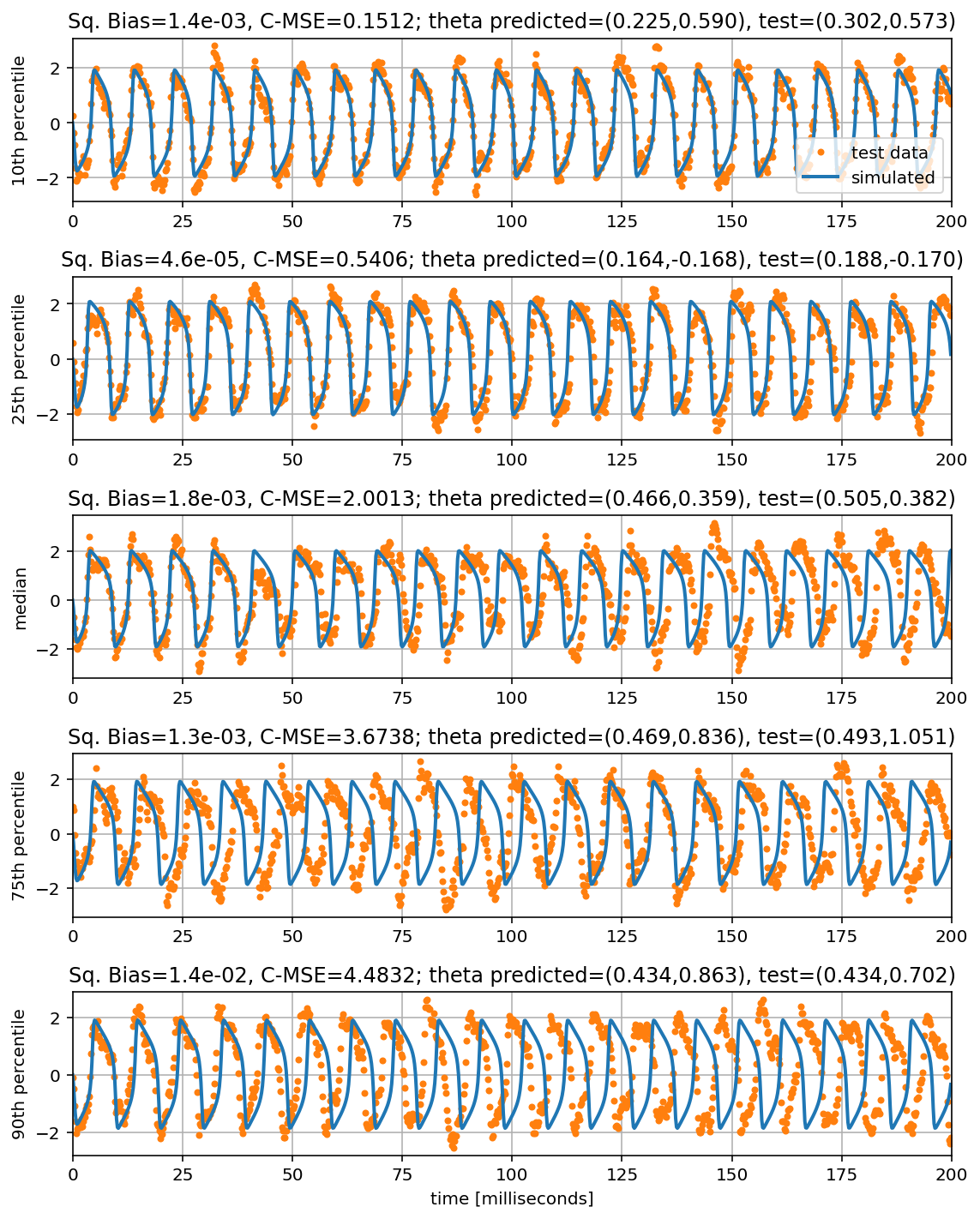}
  \caption{Simulations of FitzHugh--Nagumo model (blue lines) using parameters %
    from CNN predictions; corresponding data that gave rise to prediction are shown as orange dots.
    Training samples ($N=1000$) are noise-free but testing data ($M=2000$) contains noise.
    Graphs from top to bottom show increasing quantiles of MSE between true and estimated time series.}
  \label{fig:simODE-CNN-noiseTrain0-noiseTest1}
\end{figure}

\begin{figure}
  \centering
  \includegraphics[width=0.96\columnwidth]{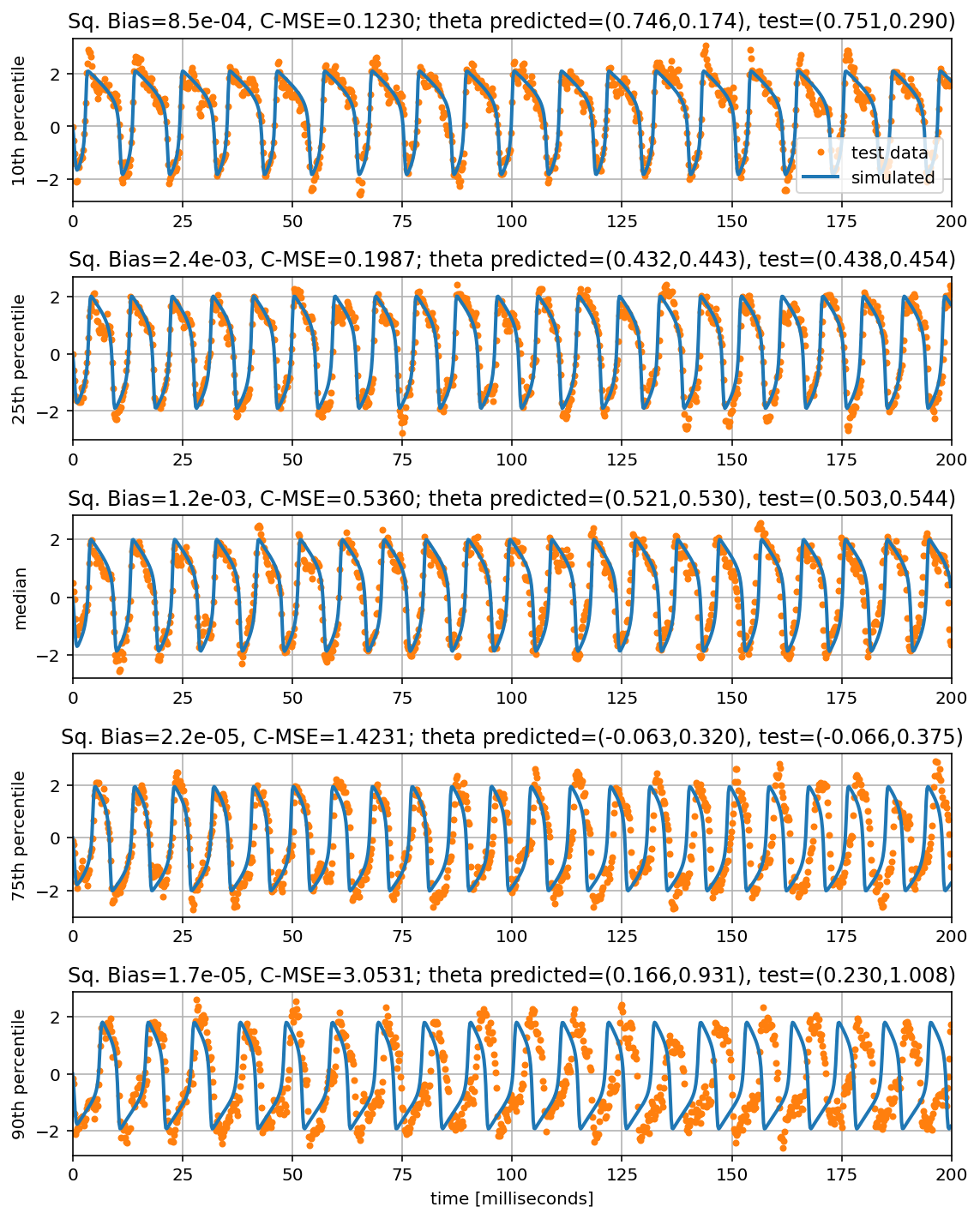}
  \caption{Simulations of FitzHugh--Nagumo model (blue lines) using parameters %
    from CNN predictions; corresponding data that gave rise to prediction are shown as orange dots.
    Training samples ($N=1000$) and testing samples ($M=2000$) both contain noise.
    Graphs from top to bottom show increasing quantiles of MSE between true and estimated time series.}
  \label{fig:simODE-CNN-noiseTrain1-noiseTest1}
\end{figure}

We show the following three Figures.  Figure~\ref{fig:simODE-CNN-noiseTrain0-noiseTest0} represents the case where both training and testing data do not contain noise; in Figure~\ref{fig:simODE-CNN-noiseTrain0-noiseTest1} only testing data contain noise; and Figure~\ref{fig:simODE-CNN-noiseTrain1-noiseTest1} shows results where both training and testing data contain noise.
Each figure contains five graphs where each graph is selected from the testing data set (of size $M=2000$) as a specific quantile of the MSE, in order to show the effects of a range of prediction accuracies on the FitzHugh--Nagumo model solutions.
We present the 10\textsuperscript{th}, 25\textsuperscript{th}, 50\textsuperscript{th} (i.e., median), 75\textsuperscript{th}, and 90\textsuperscript{th} percentiles of MSE between simulated and testing time series'.
In Figure~\ref{fig:simODE-CNN-noiseTrain0-noiseTest0}, we observe a nearly optimal overlapping of simulated output and test data for the noise-free case, even for the 90\textsuperscript{th} percentile of MSE.
When noise is added to training or testing data (Figures~\ref{fig:simODE-CNN-noiseTrain0-noiseTest1} and~\ref{fig:simODE-CNN-noiseTrain1-noiseTest1}), the simulated time series show shifted spiking frequencies, which become increasingly pronounced for the median, 75\textsuperscript{th}, and 90\textsuperscript{th} percentiles.
By comparing the predicted $\boldsymbol{\theta}=(\theta_0,\theta_1)$ with the corresponding test values (given in graph labels in Figure~\ref{fig:simODE-CNN-noiseTrain0-noiseTest1}), we observe that the prediction accuracy for $\theta_1$ degrades with increasing percentile.  This can imply that a larger contribution to discrepancies of the simulated ODE output stems from inaccurate predictions of parameter $\theta_1$.
Overall, the better correspondence of the time series' in Figure~\ref{fig:simODE-CNN-noiseTrain1-noiseTest1} compared with Figure~\ref{fig:simODE-CNN-noiseTrain0-noiseTest1} demonstrates the significant benefit of using noisy training data as it is done in the former figure.  This observation supports the results in Section~\ref{sec:results-noise}.


\clearpage
\begin{flushright}
  \framebox{\parbox{2.5in}{\scriptsize%
  \textbf{Government License:} The submitted manuscript has been created by UChicago Argonne, LLC, Operator of Argonne National Laboratory (``Argonne'').
  Argonne, a U.S. Department of Energy Office of Science laboratory, is operated under Contract No. DE-AC02-06CH11347.
  The U.S. Government retains for itself, and others acting on its behalf, a paid-up nonexclusive, irrevocable worldwide license in said article to reproduce, prepare derivative works, distribute copies to the public, and perform publicly and display publicly, by or on behalf of the Government.
  The Department of Energy will provide public access to these results of federally sponsored research in accordance with the DOE Public Access Plan. http://energy.gov/downloads/doe-public-access-plan
  }}
\end{flushright}

\end{document}